\documentclass[acmsmall]{acmart}
\usepackage{algorithm}
\usepackage{algorithmic}
\usepackage{subfigure}
\usepackage{graphics}
\usepackage{CJKutf8}
\usepackage{amsmath}
\usepackage{amssymb}
\usepackage{times}
\usepackage{soul}
\usepackage{url}
\usepackage{graphicx}
\usepackage{amsthm}
\usepackage{booktabs}
\usepackage{algorithm}
\usepackage{CJKutf8}
\usepackage{multicol}
\usepackage{multirow}
\usepackage{algorithmic}
\usepackage{bbm}
\usepackage{multibib}

\newcommand{\ie}{\emph{i.e., }}

\newcommand{\etal}{\emph{et al.}}

\AtBeginDocument{%
  \providecommand\BibTeX{{%
    \normalfont B\kern-0.5em{\scshape i\kern-0.25em b}\kern-0.8em\TeX}}}

\setcopyright{acmcopyright}
\copyrightyear{2022}
\acmYear{2022}
\acmDOI{XXXXXXX.XXXXXXX}

\acmJournal{JACM}
\acmVolume{37}
\acmNumber{4}
\acmArticle{111}
\acmMonth{8}

\begin{document}

\title{Stylized Data-to-Text Generation: A Case Study in the E-Commerce Domain}

\author{Liqiang Jing}
\email{jingliqiang6@gmail.com}
\affiliation{%
  \institution{Shandong University}
  \streetaddress{No. 72 Binhai Road, Jimo}
  \city{Qingdao}
  \state{Shandong Province}
  \postcode{266237}
    \country{China}
}

\author{Xuemeng Song}
\authornote{Corresponding author.}
\email{sxmustc@gmail.com}
\affiliation{%
  \institution{Shandong University}
  \streetaddress{No. 72 Binhai Road, Jimo}
  \city{Qingdao}
  \state{Shandong Province}
  \postcode{266237}
    \country{China}
}

\author{Xuming Lin}
\email{xuming.lxm@alibaba-inc.com}
\affiliation{%
  \institution{Alibaba Group}
  \streetaddress{No. 969 Wenyi West Road, Yuhang}
  \city{Hangzhou}
  \state{Zhejiang Province}
  \postcode{311121}
    \country{China}
}
\author{Zhongzhou Zhao}
\email{zhongzhou.zhaozz@alibaba-inc.com}
\affiliation{%
  \institution{Alibaba Group}
  \streetaddress{No. 969 Wenyi West Road, Yuhang}
  \city{Hangzhou}
  \state{Zhejiang Province}
  \postcode{311121}
    \country{China}
}
\author{Wei Zhou}
\email{fayi.zw@alibaba-inc.com}
\affiliation{%
  \institution{Alibaba Group}
  \streetaddress{No. 969 Wenyi West Road, Yuhang}
  \city{Hangzhou}
  \state{Zhejiang Province}
  \postcode{311121}
    \country{China}
}

\author{Liqiang Nie}
\email{nieliqiang@gmail.com}
\affiliation{%
  \institution{Harbin Institute of Technology (Shenzhen)}
  \streetaddress{Nanshan}
  \city{Shenzhen}
  \state{Guangdong Province}
  \postcode{518055}
    \country{China}
}








\renewcommand{\shortauthors}{L. Jing, et al.}

\begin{abstract}
Existing data-to-text generation efforts mainly focus on generating a coherent text from non-linguistic input data, such as tables and attribute-value pairs, but overlook that different application scenarios may require texts of different styles. Inspired by this, we define a new task, namely stylized data-to-text generation, whose aim is to generate coherent text for the given non-linguistic data according to a specific style. This task is non-trivial, due to three challenges: the logic of the generated text, unstructured style reference, and biased training samples. To address these challenges, we propose a novel stylized data-to-text generation model, named StyleD2T, comprising three components: logic planning-enhanced data embedding, mask-based style embedding, and unbiased stylized text generation. In the first component, we introduce a graph-guided logic planner for attribute organization to ensure the logic of generated text. In the second component, we devise feature-level mask-based style embedding to extract the essential style signal from the given unstructured style reference. In the last one, pseudo triplet augmentation is utilized to achieve unbiased text generation, and a multi-condition based confidence assignment function is designed to ensure the quality of pseudo samples. 
Extensive experiments on a newly collected dataset from Taobao\footnote{\url{https://www.taobao.com}.\label{taobal}} have been conducted, and the results show the superiority of our model over existing methods. 
\end{abstract}



\maketitle


\section{Introduction} \label{intro}
Data-to-text generation, which benefits many real-world applications, aims to convert the nonlinguistic data~\cite{DBLP:journals/jair/GattK18}, such as knowledge triples, tables, syntax trees, and attribute-value pairs, to the coherent natural language text.
One promising application of data-to-text generation is the automatic advertising article generation for products in e-commerce. Apparently, the application can not only be used to raise e-commerce platforms' sales but also reduce the labor costs of hiring professionals to manually write advertising articles for a large number of products.

Due to its great practical value, data-to-text generation has attracted increasing attention from academia and industry. Early data-to-text generation models have mainly adopted the pipeline architecture with three key modules: content planning~\cite{DBLP:conf/emnlp/DuboueM03}, sentence planning~\cite{DBLP:conf/ewnlg/DalianisH93}, and surface realization~\cite{DBLP:journals/nle/McRoyCA03}. 
With advancements in deep learning, recent studies have utilized deep neural networks and solved tasks in an end-to-end manner~\cite{DBLP:conf/emnlp/ShaoHWXZ19,DBLP:conf/coling/YuanLXWHZ20}. 
For example, 
Wiseman~\etal~\cite{DBLP:conf/emnlp/WisemanSR18} devised a hidden semi-markov model-based neural generation system, where several templates are introduced to enhance the model interpretability. To promote the diversity of the generated text,  Shao \textit{et al.}\cite{DBLP:conf/emnlp/ShaoHWXZ19} designed a planning-based hierarchical variational model with the gated recurrent units (GRUs)~\cite{DBLP:conf/emnlp/ChoMGBBSB14} to capture inter-sentence coherence and generate diversified texts.

\begin{figure}
    \centering
    \includegraphics[scale=0.55]{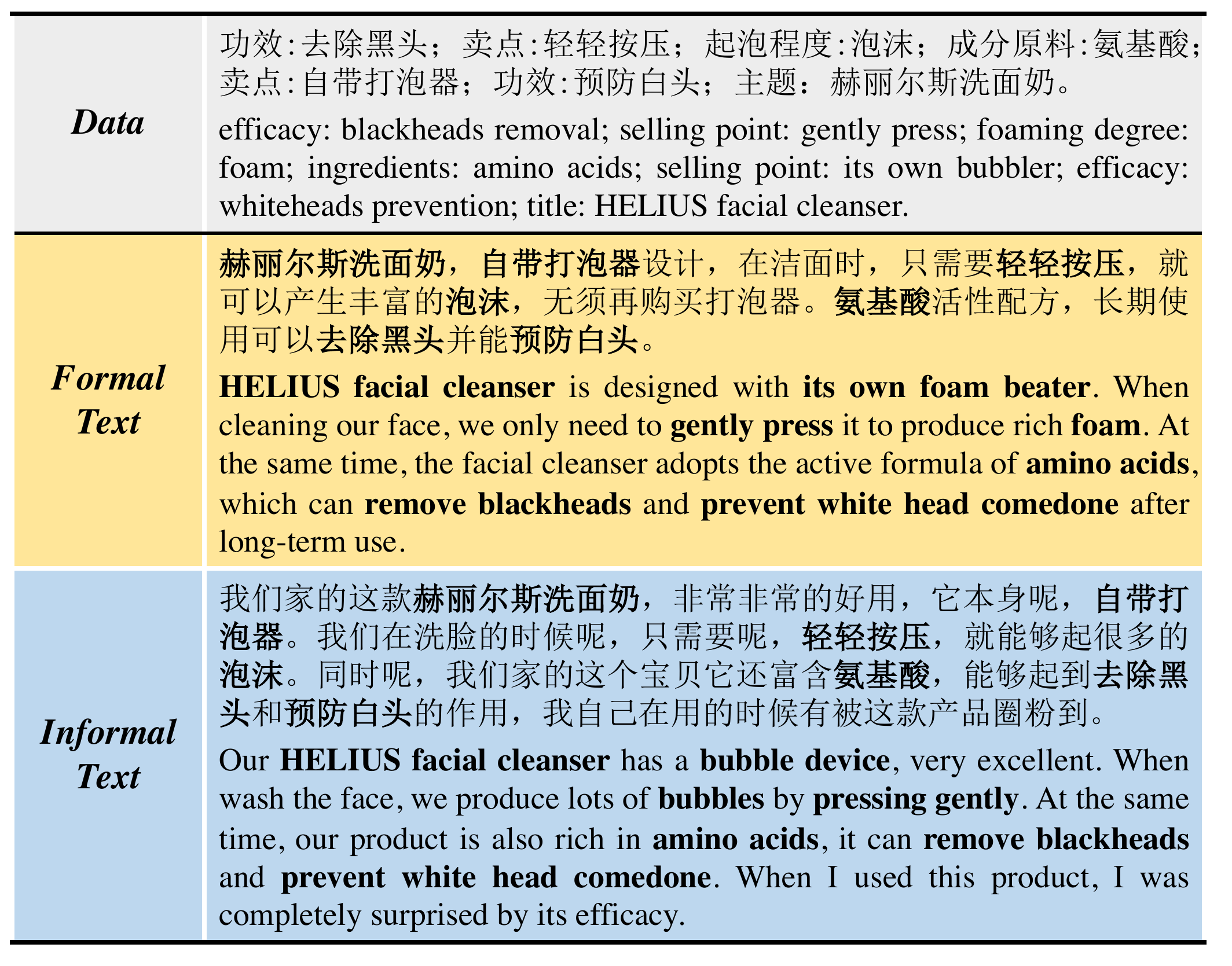}
    \caption{Illustration of stylized data-to-text generation in e-commerce, where both formal and informal advertising texts are desired. The mentioned attribute-value pairs of the given data in each text are highlighted in bold. The English texts are translated from Chinese texts, where some colloquial terms are hard to translate.}
    \label{fig:intro}
     \vspace{-0.5 em}
\end{figure}

Despite the compelling success achieved by previous efforts, these studies have mainly focused on general data-to-text generation. In other words, the linguistic quality of the generated text was emphasized as well as its semantic consistency with the given data. However, the fact that in different application scenarios, texts of different styles may be needed was overlooked.
As shown in Figure~\ref{fig:intro}, given the product data, \textit{i.e.}, a set of attribute-value pairs, formal advertising text is needed for presenting on a formal e-commerce platform, such as Weitao\footnote{\url{https://tinyurl.com/4h9fznkn}.}, but informal advertising text is desired for colloquially broadcasting in a live-streaming e-commerce scenario.

Motivated by this, we define a new task: stylized data-to-text generation, in which the goal is to generate coherent text for given nonlinguistic data according to a specific style. In a sense, the target text should maintain semantic consistency with the given data and style consistency with the specific style. As a pioneer study, we focus on nonlinguistic data in the form of attribute-value pairs, as shown in Figure~\ref{fig:intro}. To represent the desired style, instead of using fixed or learnable style embeddings, we employ a reference text of the desired style as the anchor to deliver the style information, because the linguistic style has been reported as a highly complex concept encompassing various intricate linguistic features~\cite{DBLP:conf/ijcai/YiLLS20}.

 In fact, our defined stylized data-to-text task faces three main challenges.
 \begin{itemize}
    \item \textbf{C1: The Logic of the Generated Text.}  A coherent text for the given data in terms of attribute-value pairs, as shown in Figure~\ref{fig:intro}, should maintain a logical flow, \textit{i.e.}, the attributes of the data should be described in a certain logical order, making it natural and easy to interpret.  Consequently, guaranteeing the logic of the generated text is a key challenge.
    \item \textbf{C2: Unstructured Style Reference.} The given style reference text is unstructured and expresses both the semantic and style information. Accordingly, how to accurately derive the style information without the semantic content from the unstructured style reference text is another crucial challenge. 
    \item \textbf{C3: Biased Training Samples.} It is unrealistic to simultaneously obtain the corresponding text for given data for each style (\textit{e.g.,} formal and informal styles). Thus, optimizing the model with biased training samples may lead to the biased generation of text, \textit{i.e.}, the stylized text generator may be misled and generate the target text based only on the given data without considering the style information. Therefore, how to achieve unbiased stylized text generation is a vital challenge.
\end{itemize}

\begin{figure*}[!t]
  \centering
  \includegraphics[scale=0.52]{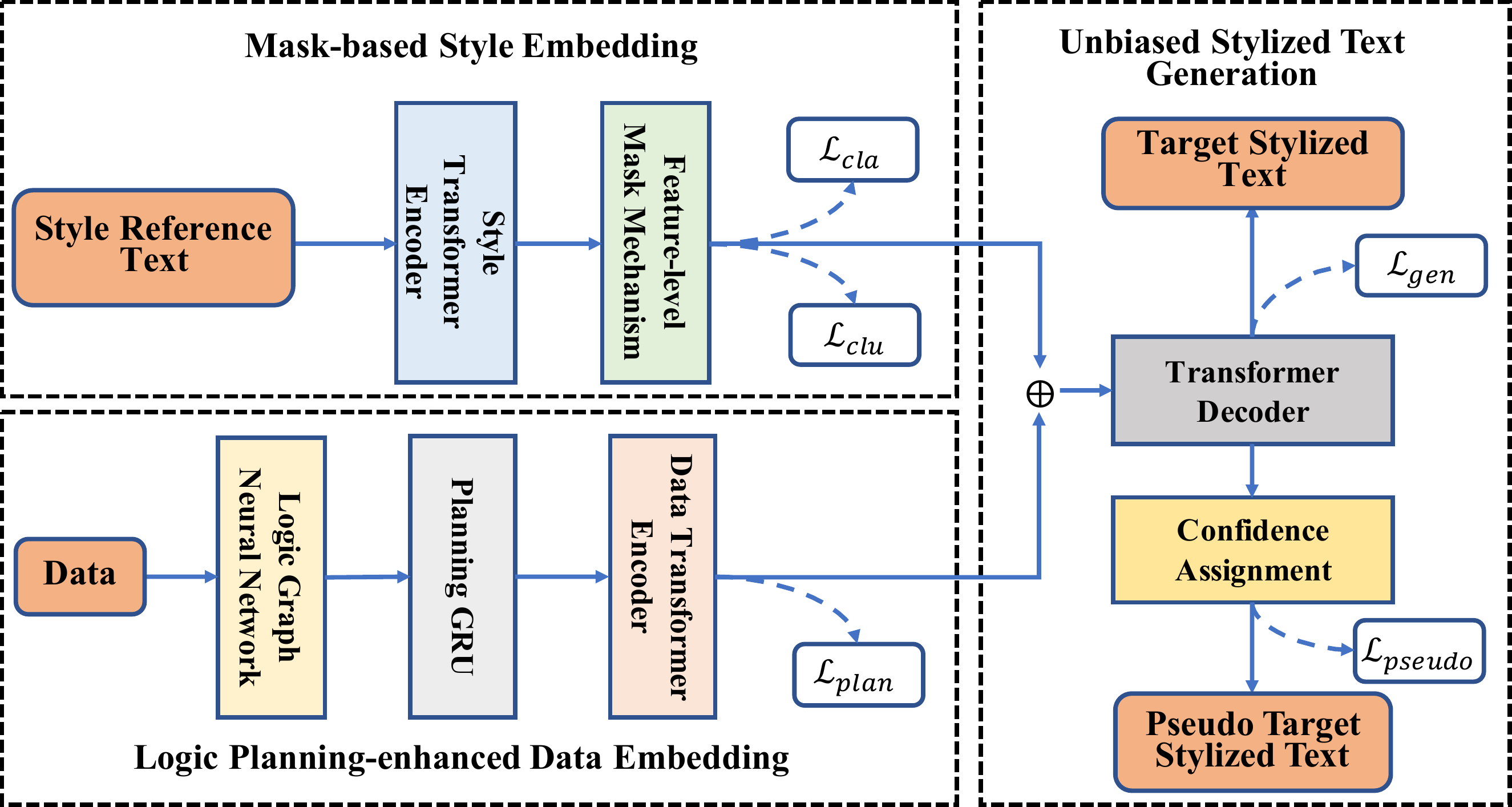}
  \caption{Illustration of the proposed scheme StyleD2T, which consists of three components: logic planning-enhanced data embedding, mask-based style embedding, and unbiased stylized text generation. The concatenation operation is denoted as ``$\bigoplus$''.}
  \label{framework}
  \vspace{-0.5 em}
\end{figure*}

To tackle these challenges, we propose a novel stylized data-to-text generation model for product advertising generation, StyleD2T. As shown in Figure~\ref{framework}, StyleD2T consists of three components, namely, logic planning-enhanced data embedding, mask-based style embedding, and unbiased stylized text generation.
In particular, for \textbf{C1}, we  design a graph-guided logic planner in the first component, where a logic graph is introduced to characterize the logical relation among attribute-value pairs of the given data of the product and plan the attribute organization for generating the target text with the GRU. We also introduce the data transformer encoder to obtain the context embedding of the planned attribute.
Based on the concern that some words may simultaneously contain style and content information, to tackle \textbf{C2}, we propose the feature-level mask-based style embedding component, as opposed to adopting high-level or low-level masks as in existing methods.
Regarding \textbf{C3}, we introduce pseudo triplet augmentation in the third component, whereby a multi-condition-based confidence assignment function is devised to ensure the quality of pseudo samples.
To verify the effectiveness and generality of our model, we collect a real-world dataset named TaoStyle, from Taobao, a very large Chinese online shopping site. This dataset supports stylized data-to-text generation for the informal and formal styles. Our model has yielded superior performance compared with state-of-the-art methods on this real-world dataset.

Our contributions can be summarized as follows:

\begin{itemize}
    \item We define a new research task, stylized data-to-text generation, which is motivated by realistic application scenarios that indeed require different writing styles. This new task emphasizes that the generated text should be not only coherent and semantically consistent with the given data but also written in a manner that conforms to the desired style.
    \item We propose a novel stylized data-to-text generation framework comprising three key components: logic planning-enhanced data embedding, mask-based style embedding, and unbiased stylized text generation, which can effectively fulfill the proposed task.
    \item To verify our StyleD2T model, we construct a real-world dataset from the popular Chinese e-commerce platform Taobao. Extensive experiments on this dataset show the effectiveness of our model. In addition, the codes have been released as a byproduct\footnote{\url{https://github.com/LiqiangJing/StyleD2T}}.
\end{itemize}

The remainder of this article is as follows. We first briefly review the related work in Section~\ref{sec:rel}. Next, we formulate the proposed task and detail our StyleD2T in Section~\ref{sec:method}. Then, the experimental setup and the analysis of the results are presented in Section~\ref{sec:exp}. Finally, the conclusion and future research directions are described in Section~\ref{sec:conclusion}.

\section{Related Work} \label{sec:rel}
Our work is related to the studies of data-to-text generation, stylized generation, graph neural networks, and transformer-based models for neural language processing.

\subsection{Data-to-Text Generation} 
Early data-to-text generation methods~\cite{DBLP:conf/inlg/McRoyCA00,DBLP:conf/acl/KondadadiHS13} mainly adopted the pipeline architecture with separate components, such as content planning, sentence planning, and surface realization. Content planning aims to determine what information should appear in the generated text and the structure of the text. For example, Duboue \textit{et al.}~\cite{DBLP:conf/emnlp/DuboueM03} proposed a method to automatically acquire the content selection rules from a corpus of text and its associated semantics.
Sentence planning is used to determine the structure and lexical content of each sentence in the generated text. For example, Daliants \textit{et al.}~\cite{DBLP:conf/ewnlg/DalianisH93} presented eight aggregation strategies to avoid repetition of the selected material during text generation.
The aim of surface realization is to convert the sentence plan to a natural language string, which can be classified into template-based and grammar-based approaches. Template-based approaches~\cite{DBLP:journals/nle/McRoyCA03,DBLP:journals/coling/DeemterTK05} utilize templates defined by knowledge experts to convert the sentence plan to the text. In contrast, in grammar-based approaches~\cite{DBLP:journals/nle/Bateman97,DBLP:conf/acl/EspinosaWM08}, a grammatical system to construct surface strings is usually built.

Recently, with the evolution of deep neural networks, several end-to-end data-to-text generation models have emerged. For example, Jain \textit{et al.}~\cite{DBLP:conf/naacl/JainLSNKS18} proposed an end-to-end mixed hierarchical attention-based encoder-decoder model to generate summaries of table data, while Nema \textit{et al.}~\cite{DBLP:conf/naacl/NemaSJLSK18} introduced an end-to-end fused bifocal attention model with a gated orthogonalization mechanism for exploiting task-specific characteristics. 
To enhance the content fidelity of the generated texts with the input attribute-based data, Yuan \etal \cite{DBLP:conf/coling/YuanLXWHZ20} devised a dual-copy mechanism, which copies tokens from the textual product description and product attributes to the generated text. To deal with a large vocabulary, Lebret \textit{et al.}~\cite{DBLP:conf/emnlp/LebretGA16} extended the conditional neural language model to mix a fixed vocabulary with the copy mechanism so that sample-specific words can be transferred from the input database to the generated sentence. Moreover, to promote the logic of the generated text,  Sha \textit{et al.}~\cite{DBLP:conf/aaai/ShaMLPLCS18} devised a table field linking mechanism to capture the relationship between different fields in the table and used this relationship to enable the model to better plan what to output first and what to output next.
Noticed the practical demand of diversified expression, Shao \textit{et al.}~\cite{DBLP:conf/emnlp/ShaoHWXZ19} proposed an end-to-end planning-based hierarchical variational approach to diversify the content plan and hence generate diversified texts.
In addition, Ye \textit{et al.}~\cite{DBLP:conf/iclr/YeS0W020} proposed a variational template machine, which utilizes diverse templates to generate diverse texts.

Although existing methods have achieved great success, their focus was on general data-to-text generation. The fact that generated text should be in accordance with certain styles in many application scenarios was overlooked, which is the major concern of this work. 

\subsection{
Stylized Generation
} 
Stylized generation refers to the generation of text with a specified style. Because stylized generation can be applied to various application scenarios, it has been explored in multiple research tasks.
For example, some studies have explored the classic stylized generation task called text style transfer, in which the aim is to generate new text by changing the style of
the original text while keeping its content unchanged. Specifically, existing studies of text style transfer can be roughly divided into two categories: disentanglement approaches and attention-based approaches. In the former approaches, the content and style from the original text are explicitly disentangled and then separated content and style representations are incorporated to generate the target text~\cite{DBLP:conf/nips/ShenLBJ17,DBLP:conf/acl/JohnMBV19}. In the latter approaches, a trainable or fixed style embedding is utilized as a style signal and the model is encouraged to focus on style-independent words~\cite{DBLP:conf/acl/DaiLQH19,DBLP:conf/acl/HuangCWGZH21}.
With respect to dialogue systems, Zhou \textit{et al.}~\cite{DBLP:conf/aaai/ZhouHZZL18} proposed an emotional chatting machine that can generate responses with a specified emotion. Later, Kong \textit{et al.}~\cite{DBLP:conf/acl/KongHTGH21} introduced the task of stylized story generation, in which the goal is to generate a coherent story with a specified style given the first sentence as the leading context in story generation.

Inspired by these efforts, we define a new task of stylized data-to-text generation, which differs from conventional data-to-text generation by requiring the generated text to have a specific style. This task can be applied to more scenes compared to general data-to-text generation.

\subsection{Graph Convolutional Networks}
Recently, studies on graph convolutional networks (GCNs) have received increasing attention because of their superior ability to model structured data, \textit{i.e.}, graphs~\cite{graph1,graph2,graph3,graph4}. The original GCN algorithm proposed by Kipf \textit{et al.}~\cite{DBLP:conf/iclr/KipfW17} learns hidden representations of nodes by aggregating neighbor information. Due to the great power of the GCN, it has been widely applied in multiple fields, including natural language processing~(NLP), computer vision, and recommendation systems.
For example, in the NLP domain, Chen~\etal~\cite{DBLP:journals/tois/ChenSRZCN20} devised a fine-grained privacy detection network that explored the semantic correlations among personal aspects with a GCN.
In addition, in the computer vision domain,
Caramalau~\textit{et al.}~\cite{DBLP:conf/cvpr/CaramalauBK21} presented a novel sequential GCN to learn node representations and distinguish sufficiently different unlabeled examples from labeled examples for active learning, and Zhang \etal \cite{DBLP:journals/tip/ZhangHSYN21} devised a multimodal interaction GCN to jointly explore the complex intra-modal relations and inter-modal interactions for temporal language localization in videos. 
Zheng~\etal~\cite{DBLP:conf/mm/ZhengSNDZN21} integrated  disentangled item representations into a GCN to adaptively propagate the fine-grained compatibility relationships among items for outfit compatibility modeling,
Wang~\textit{et al.}~\cite{DBLP:conf/sigir/Wang0WFC19} developed a novel neural graph collaborative filtering method that integrated user-item interactions into a user embedding process for recommendation systems.

According to the abovementioned studies, the GCN has demonstrated its superiority in learning the structure information among nodes in the graph, which inspires us to utilize the GCN for logic correlation modeling among attribute-value pairs of the given input.

\subsection{Transformer-based Models for Natural Language Processing}
The transformer~\cite{DBLP:conf/nips/VaswaniSPUJGKP17} model, which consists of an encoder and a decoder, was initially proposed for machine translation.
The encoder learns interactions between the left and right context words with the bidirectional attention mechanism, while the decoder is designed to generate the target translation text using the unidirectional attention mechanism.
Due to the remarkable capability of attention mechanisms, many transformer-based models have emerged in the NLP domain.
According to the architecture, current transformer-based models for NLP can be roughly classified into three groups: transformer encoder-based models, transformer decoder-based models, and full transformer-based models~\cite{DBLP:journals/corr/abs-2003-08271}. 
The typical transformer encoder-based models are BERT~\cite{DBLP:conf/naacl/DevlinCLT19} and RoBERTa~\cite{DBLP:journals/corr/abs-1907-11692}, both of which adopt the transformer encoder as the backbone and are trained by the masked language modeling technique. 
Regarding the transformer decoder-based model, a classic example is GPT~\cite{gpt}, which is developed based on  the transformer decoder and achieves comparable performance with the above two models (\ie BERT and RoBERTa) on many natural language understanding tasks. 
Regarding the full transformer-based model,  BART~\cite{DBLP:conf/acl/LewisLGGMLSZ20} and T5~\cite{DBLP:journals/jmlr/RaffelSRLNMZLL20} are two representative models that were proposed for tackling natural language generation tasks.
 
To date, these transformer-based models have achieved compelling success in various NLP tasks, such as movie script generation~\cite{transformer1}, personalized answer generation~\cite{transformer2}, and query suggestion~\cite{transformer3}.
Inspired by their remarkable performance, we utilize the transformer encoder and decoder for our text representation and generation, respectively.


\begin{center}
\begin{table*}[]
\centering
\setlength{\abovecaptionskip}{0cm}
\setlength{\belowcaptionskip}{0cm}
\caption{\upshape Summary of the key notations.}
\label{notations}
\setlength{\tabcolsep}{1.2mm}{
\begin{tabular}{c|c}
\hline
\textbf{Symbol} & \textbf{Explanation} \\
 \hline \hline
$\mathcal{D}$ & The set of training triplets. \\
$\mathcal{P}_j$ & The input data (\ie a set of attribute-value pairs) of the $j$-th triplet. \\
$p_j^k$ & The $k$-th  attribute-value pair of the input data $\mathcal{P}_j$. \\
$X_j$ & The style reference text of the $j$-th  triplet. \\
$Y_j$ & The target text of the $j$-th  triplet.\\
$\hat{Y}_j$  & The generated target text of the $j$-th triplet.\\
$N$ & The total number of training triplets.\\
$N_s$ & The total number of styles.\\
$\textbf{g}_j$ & The ground-truth one-hot style label vector of the $j$-th  triplet.\\
$\hat{\textbf{g}}_j$ & The predicted style vector of the $j$-th  triplet.\\
$\textbf{E}^k$ & The token embedding matrix of the $k$-th attribute-value pair.  \\
$\mathcal{C}$ & The corpus containing all the target stylized texts in the training dataset.\\
$r_i^Y$ & The ground-truth ranking of the attribute-value pair $p^i$ in the text $Y$.\\
$L$ & The ground-truth attribute-value pair organization plan.\\
$\hat{L}$ & The predicted attribute-value pair organization plan.\\
$\mathbf{H}_o$ & The planning-enhanced representation for the input data.\\
$\mathbf{E}_X$ & The embedding of the style reference text $X$.\\
$\mathbf{H}_X$ & The encoded representation of the style reference text $X$.\\
$\mathbf{M}$ & The feature-level mask matrix.\\
$\mathbf{s}$ & The style embedding derived from the style reference text $X$.\\
$\alpha$, $\beta$, $\gamma$, $\delta$ &  The hyperparameters for loss functions.\\

\hline
\end{tabular}
}
\end{table*}
\end{center}

\section{Method} \label{sec:method}

In this section, the problem formulation is first introduced, and then the three components of our proposed StyleD2T model are detailed. Finally, the training and inference processes are presented.

\subsection{Problem Formulation \label{problem}}
Suppose that we have a set of $N$ training triplets $\mathcal{D}=\{(\mathcal{P}_1, X_1, Y_1), (\mathcal{P}_2, X_2, Y_2), \cdots, (\mathcal{P}_N,  X_N, Y_N)\}$.
$\mathcal{P}_j=\{p_j^1, p_j^2, \cdots, p^{K_j}_j\}, j=1,\cdots, N$, is the input data of the $j$-th triplet, which is represented by a set of $K_j$ attribute-value pairs, where $p_j^k$ is the $k$-th attribute-value pair, $k=1,2,\cdots, K_j$. $p_j^k=\{w_j^{k,1}, w_j^{k,2}, \cdots, w_j^{k,m}\}$ and $w_j^{k,i}$ is the $i$-th token in $p_j^k$. 
$X_j$ is the style reference text of the $j$-th triplet, which shares the same style as the target stylized text $Y_j$.
In particular, $N_s$ refers to the number of possible styles that we can specify.
To facilitate style embedding learning, we assign each $X_j$ a one-hot label vector $\mathbf{g}_j=[g_j^1, g_j^2,\cdots,g_j^{N_s}]\in \{0,1\}^{N_s}$, where $g_j^q=1$ if $X_j$ belongs to the $q$-th style; otherwise, $g_j^q=0$.
$Y_j$ is the target text of the $j$-th triplet, which meets the content fidelity with the given data $\mathcal{P}_j$ and the style consistency with the style reference text $X_j$.
Formally, we use $\mathcal{C}=\{ Y_1, Y_2, \cdots, Y_N\}$ to denote the whole corpus.  Table~\ref{notations} summarizes the key notations.
Ultimately, our goal is to learn a model $\mathcal{M}$ that is able to take $(\mathcal{P}, X)$ as the input and generate the target stylized text $\hat{Y}$ as follows:
\begin{equation}
\label{formulation}
    \hat{Y} = \mathcal{M}(\mathcal{P}, X |\boldsymbol{\Theta}),
\end{equation}
where $\boldsymbol{\Theta}$ refers to the parameters to be learned. 
For simplicity, we temporarily omit the index (\textit{i.e.}, the subscript $j$) of each training triplet.

\begin{figure*}[!t]
  \centering
  \includegraphics[scale=0.5]{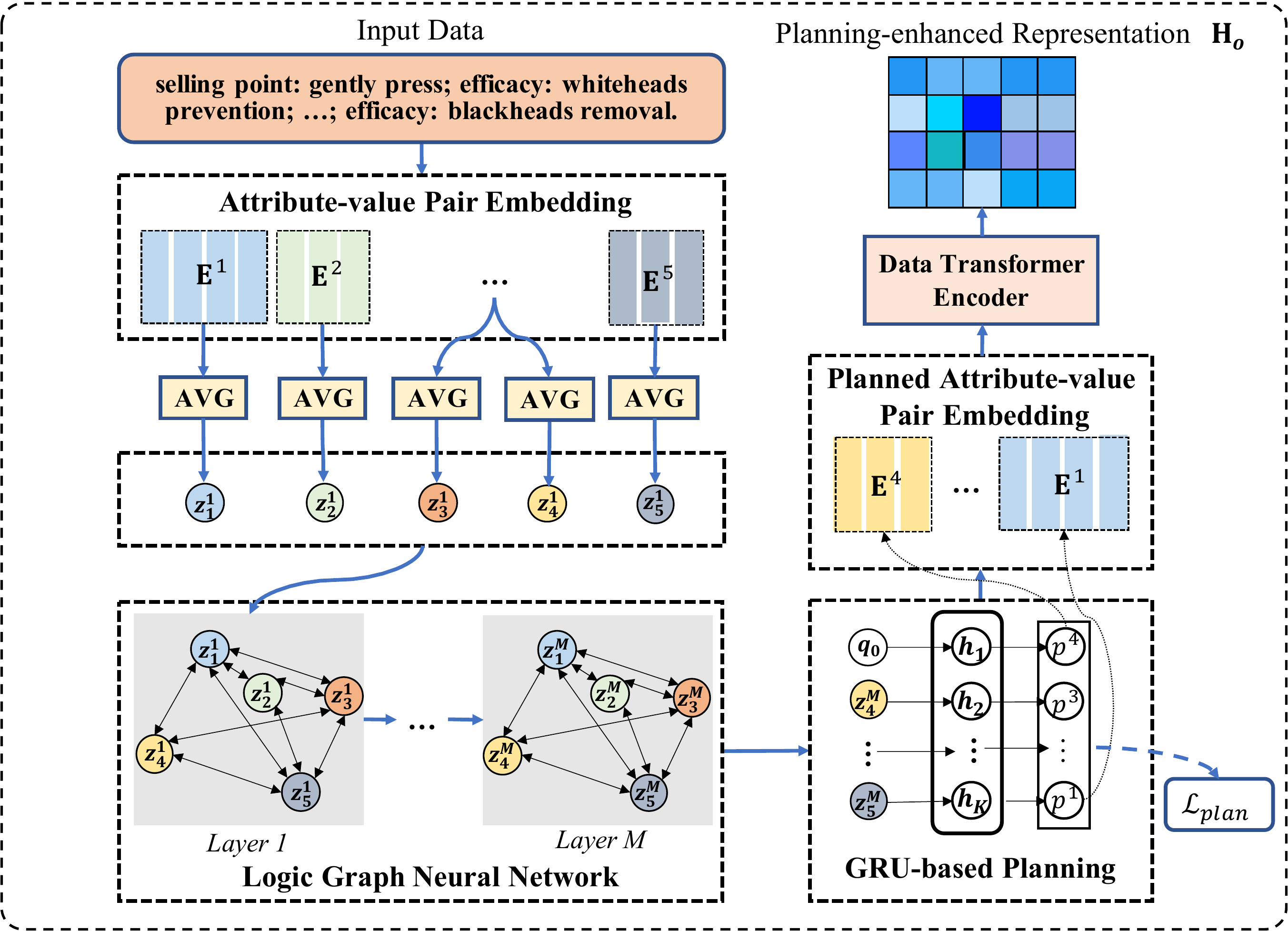}
  \caption{Illustration of the logic planning-enhanced data embedding component, where $K=5$.}
  \label{fig:planner}
   \vspace{-0.5em}
\end{figure*}
\subsection{ Logic Planning-enhanced Data Embedding\label{dataemb}}
As shown in Figure~\ref{fig:planner}, the logic planning-enhanced data embedding component works on the data embedding, \textit{i.e.}, the attribute-value pair embedding, and organizes the attribute-value pairs in the target text.

\textbf{Attribute-value Pair Embedding.} Suppose that the input data $\mathcal{P}=\{p^1,p^2,\cdots,p^K\}$ is represented by $K$ attribute-value pairs. We first embed each attribute-value pair with the embedding layer of the pretrained language model BART~\cite{DBLP:conf/acl/LewisLGGMLSZ20} as, 
\begin{equation}
\label{embedding}
    \mathbf{E}^k= BART\_Embedding(p^k), 
\end{equation}
where $\mathbf{E}^k\in \mathbb{R}^{t_{k} \times d_1}$ refers to the token embedding matrix of the $k$-th attribute-value pair of the data $\mathcal{P}$, whose $t$-th row refers to the embedding of the $t$-th token of the attribute-value pair $p^k$. $t_{k}$ is the total number of tokens in $p^k$, and $d_1$ is the dimension of token embedding. 

\textbf{Graph-guided Logic Planner.}
Since each input data has multiple attribute-value pairs, we need to generate a long text (\textit{i.e.},  multiple sentences) to comprehensively describe it. 
One way to generate long text for data that involves a set of attribute-value pairs is to learn how to arrange the order of the attribute-value pairs to be described. 
To ensure that the transition flow of the generated long text is logical, we devise a graph-guided logic planner, which 1) first builds a logic graph for each data to characterize the logical relation among all attribute-value pairs, 2) then employs the GCN~\cite{DBLP:conf/iclr/KipfW17} to refine the embedding of each attribute-value pair by absorbing its logical context information, and 3) finally adopts the GRU~\cite{DBLP:conf/emnlp/ChoMGBBSB14} to fulfill the attribute-value pair planning.

\begin{figure*}[!t]
  \centering
  \includegraphics[scale=0.46]{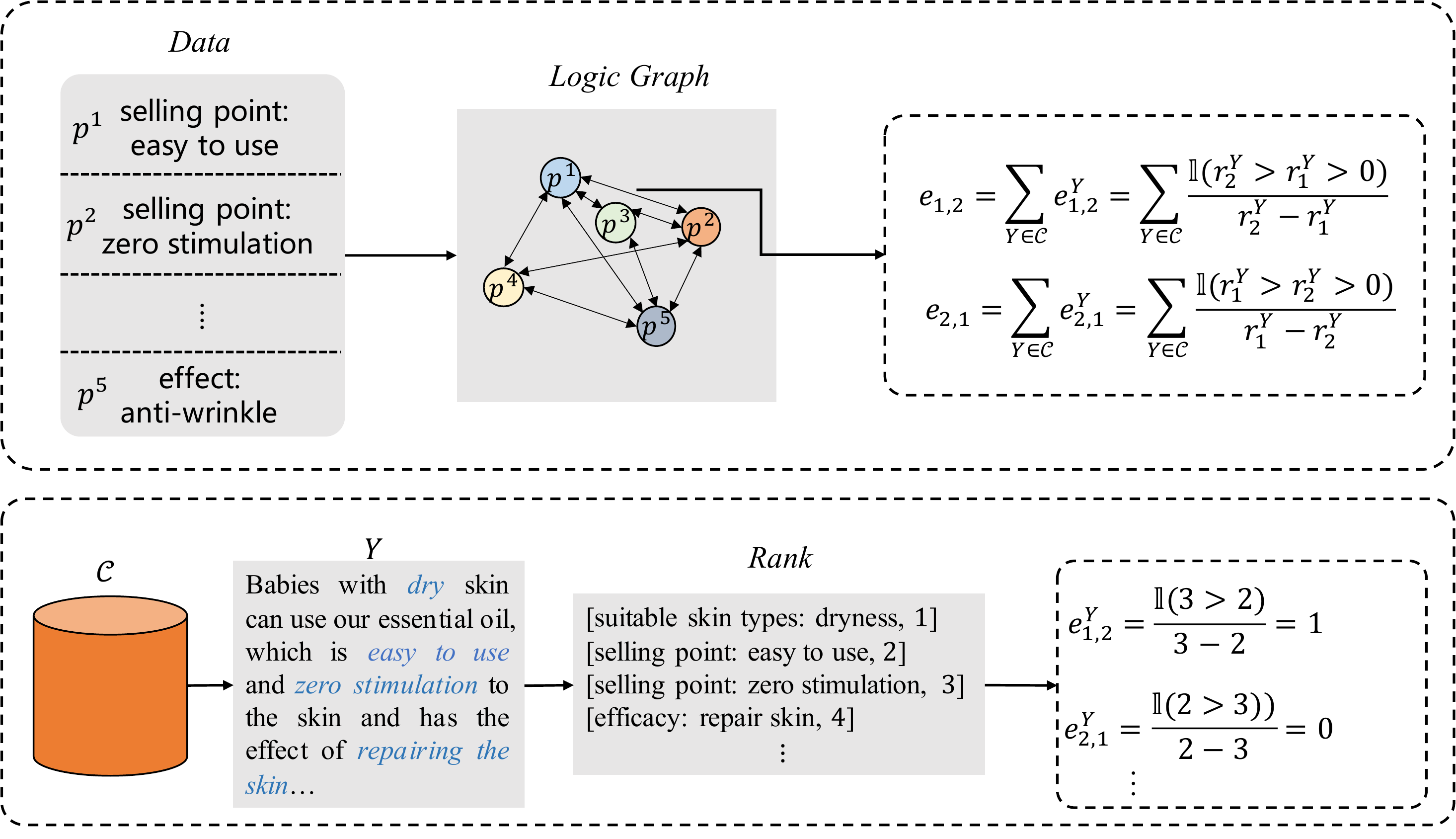}
  \caption{Illustration of the logic graph construction. $\mathcal{C}$ denotes all the stylized texts existing in the training set. $Y$ is a stylized text. The top part is the instance of computing edges between node ``selling point: easy to use'' (\ie $p^1$) and node ``selling point: zero 
simulation
'' (\ie $p^2$). The bottom part shows the computation of edges $e_{1,2}^Y$ and $e_{2,1}^Y$ for the stylized text $Y$ existing in $\mathcal{C}$.} 
  \label{fig:logic_graph}
   \vspace{-0.5 em}
\end{figure*}

\textit{Logic Graph Construction.} To comprehensively capture the logical relation among all attribute-value pairs of the given data, we build a graph $\mathcal{G}=\{\mathcal{E}, \mathcal{R}\}$, where $\mathcal{E}=\{n_1,n_2,\cdots,n_K\}$ denotes the set of nodes, and node $n_i$ corresponds to the $i$-th attribute-value pair of the given data, \textit{i.e.}, $p^i$. 
Each node $n_i$ is associated with a hidden vector $\mathbf{z}_i$, which is updated at each iteration of the information propagation over the graph. 
$\mathcal{R}=\{(n_i,n_j)|i, j \in [1,\cdots, K]\}$ represents the set of edges among these nodes, reflecting the logical relation among all attribute-value pairs. In particular, we utilize the statistics of the relative positions of each two attribute-value pairs in texts of our corpus $\mathcal{C}=\{Y_1,\cdots,Y_N\}$ to indicate their logical relations. Specifically, we assign each edge $(n_i, n_j)$ a weight $e_{i,j}$, which is defined as follows:
\begin{equation}
\label{graph_cons}
    \begin{aligned}
     e_{i,j} =
     &\sum_{Y \in \mathcal{C}} \mathbbm{1}(r^{Y}_{j} > r^{Y}_{i}>0)/ ({r^{Y}_{j} - r^Y_i}), 
    \end{aligned}
\end{equation}
where $\mathbbm{1}(\cdot)$ is the indicator function. $r^{Y}_i\in \{0,1, \cdots,K\}$ indicates that the ranking of the attribute-value pair $p^{i}$ appears in the text $Y$. Notably, $r^{Y}_i=0$ when the attribute-value pair $p^{i}$ fails to appear in the text $Y$. For example, the rankings for the attribute-value pair ``selling point: its own bubbler'' and ``selling point: gently press'' in the formal text shown in Figure~\ref{fig:intro} are $1$ and $2$, respectively.
Intuitively, the attribute-value pairs with the closer relative positions, \textit{i.e.}, the smaller $(r^{Y}_j - r^{Y}_i)$, have the stronger logical relation. 
An instance of a logic graph is shown in Figure~\ref{fig:logic_graph}.

\textit{Logical Context Propagation.} Due to its remarkable performance in nonsequential data representation learning, we adopt the GCN to refine each attribute-value pair embedding by logical context propagation as follows:
\begin{equation}
\label{graph}
\left \{
    \begin{aligned}
    \mathbf{z}_i^{l+1} &= \eta\Big[\mathbf{W}_c^{l} \Big(\sum_{n_j \in \mathcal{N}_{i}}\frac{e_{i,j} \mathbf{W}_a^{l} \mathbf{z}_j^{l}}{e_{i,*}} 
    + \mathbf{W}_b^{l} \mathbf{z}_{i}^l\Big)+ \mathbf{b}_c^{l}\Big], 
    \\e_{i,*} &= \sum_{n_{j^\prime} \in \mathcal{N}_{i}} e_{i, j^\prime},
    \end{aligned} 
\right.
\end{equation}
where $\mathcal{N}_{i}$ is the set of neighbor nodes of node $n_i$. $\mathbf{W}_a^{l}\in \mathbb{R}^{d_1 \times d_1}$, $\textbf{W}_b^{l}\in \mathbb{R}^{d_1 \times d_1}$, $\mathbf{W}_c^{l} \in \mathbb{R}^{d_1 \times d_1}$ and $\mathbf{b}_c^{l} \in \mathbb{R}^{d_1}$ are the learnable parameters for the $l$-th propagation layer, where $l=1,2,\cdots,M$. $M$ is the total number of propagation layers. 
$\eta$ denotes the tanh activation function.
$\mathbf{z}_j^{l}$ is the input hidden state vector for node $n_j$ of the $l$-th layer, where we set $\mathbf{z}_i^1=AVG(\mathbf{E}^i)$. $AVG(\cdot)$ is the average pooling operation. Finally, we regard the last layer output of the GCN (\textit{i.e.}, $\mathbf{z}_i^{M+1}$) as the refined embedding of the $i$-th attribute-value pair, which is denoted as $\mathbf{a}_i$. 


\textit{GRU-based Planning.}
Intuitively, since we can derive the ground-truth plan from the ground-truth stylized text $Y$, we can directly use this plan to reorganize the refined embeddings of all the attribute-value pairs of the given data. In particular, we employ the GRU to learn to logically organize the attribute-value pairs based on the refined attribute-value pair embeddings that have absorbed the logical context. 
Let $L=[L_1, L_2, \cdots, L_K]=[p^{m_1},p^{m_2},\cdots,p^{m_K}]$ denote the ground-truth attribute-value pair organization plan for the given data $\mathcal{P}$, which can be derived from the ground-truth target text $Y$. 
$L_t=p^{m_t}$ refers to the $t$-th mentioned attribute-value pair in the text $Y$, where $m_t\in[1,K]$ is the original index of $L_t$ in the data $\mathcal{P}$. We then arrange the attribute-value pairs of $\mathcal{P}$ in a sequence according to $L$ and feed them into the GRU as follows:
\begin{align}
\label{gru}
    \textbf{o}_t,\textbf{h}_t = GRU(\mathbf{q}_{t-1}, \textbf{h}_{t-1}),
\end{align}
where $\mathbf{h}_t\ \in \mathbb{R}^{d_1}$ is the hidden state vector at step $t$ that encodes the historical logical context. $\mathbf{o}_t\ \in \mathbb{R}^{d_1}$ is the output of the GRU that is utilized to predict the $t$-th attribute-value pair. 
The first hidden state is initialized by the average of all the refined node embeddings, \textit{i.e.}, $\textbf{h}_0=AVG(\mathbf{{a}}_1,\cdots,\mathbf{{a}}_K )$.
$\textbf{q}_{t-1}={\textbf{a}}_{m_{t-1}}$ is the input vector for predicting the $t$-th attribute-value pair and refers to the refined logical embedding of the $(t-1)$-th attribute-value pair described in the ground-truth stylized text $Y$ when $t=2,\cdots, K$.
$\mathbf{q}_{0}$ is a learnable parameter for predicting the first attribute-value pair $p^{m_1}$ to be described.
Finally, given $\mathcal{P}$, we can define the probability of the ground-truth plan $L$ as follows:
\begin{equation}
\label{eq6}
\left \{
    \begin{aligned}
    &P(L|\mathcal{P}) = \prod_{t=1}^{K} P(L_t|L_{<t}, \mathcal{P}),\\
    &P(L_t|L_{<t}, \mathcal{P})=softmax_{L_t}(\mathbf{o}_t^\top\mathbf{W}_L \mathbf{A}),
    \end{aligned} 
\right.
\end{equation}
where $L_t$ and $L_{<t}$ are the $t$-th attribute-value pair and all its previous attribute-value pairs in $L$, respectively. $P(L_t|L_{<t}, \mathcal{P})$ refers to the probability that the $t$-th predicted attribute-value pair is $L_t$ given the previous $t-1$ attribute-value pairs of plan $L$. $\mathbf{A}=[\mathbf{{a}}_{m_1};\cdots; \mathbf{{a}}_{m_{K}}] \in \mathbb{R}^{d_1 \times m_K}$, where $[;]$ is the concatenation operation. $\mathbf{W}_L \in \mathbb{R}^{d_1 \times d_1}$ is a mapping matrix to be learned, and $softmax_{L_t}(\cdot)$ denotes the $t$-th element of the softmax normalized predicted probability distribution over all possible attribute-value pairs. 
Finally, in order to ensure correct planning for the given data, we use the following loss function:
\begin{equation}
\label{eq7}
    \mathcal{L}_{planning}= -\log {P(L|\mathcal{P})}.
\end{equation}

\textbf{Planning-enhanced Data Representation.} To utilize the logic relationship existing in the input data, we directly use the plan predicted by the graph-guided logic planner rather than the original data to obtain a planning-enhanced data representation.
Specifically, we first obtain the planned attribute-value pair embeddings of the training data $\mathcal{P}$ according to the predicted plan $\hat{L}=[p^{\hat{m_1}},p^{\hat{m_2}},\cdots,p^{\hat{m_K}}]$ and then derive the data's planning-enhanced representation with a transformer encoder~\cite{DBLP:conf/nips/VaswaniSPUJGKP17} 
due to its great success in many NLP tasks~\cite{DBLP:conf/ijcai/LiTZW21}. Formally, we have
\begin{equation} \label{planning}
 \begin{aligned}
\mathbf{H}_o=Transformer_o([\textbf{E}^{\hat{m_1}};\cdots;\textbf{E}^{\hat{m_K}}]), 
\end{aligned}
 \end{equation}
where  $\mathbf{E}^{\hat{m_k}}$ denotes the token embedding matrix of the $k$-th attribute-value pair in the predicted plan, which is obtained by Eqn. ($\ref{embedding}$). 
\mbox{$\mathbf{H}_o \in \mathbb{R}^{T \times d_1}$} is the planning-enhanced representation for the data $\mathcal{P}$, and $T=\sum_{k=1}^K t_k$ is the total number of tokens of $\mathcal{P}$.

\begin{figure*}[!t]
  \centering
  \includegraphics[scale=0.55]{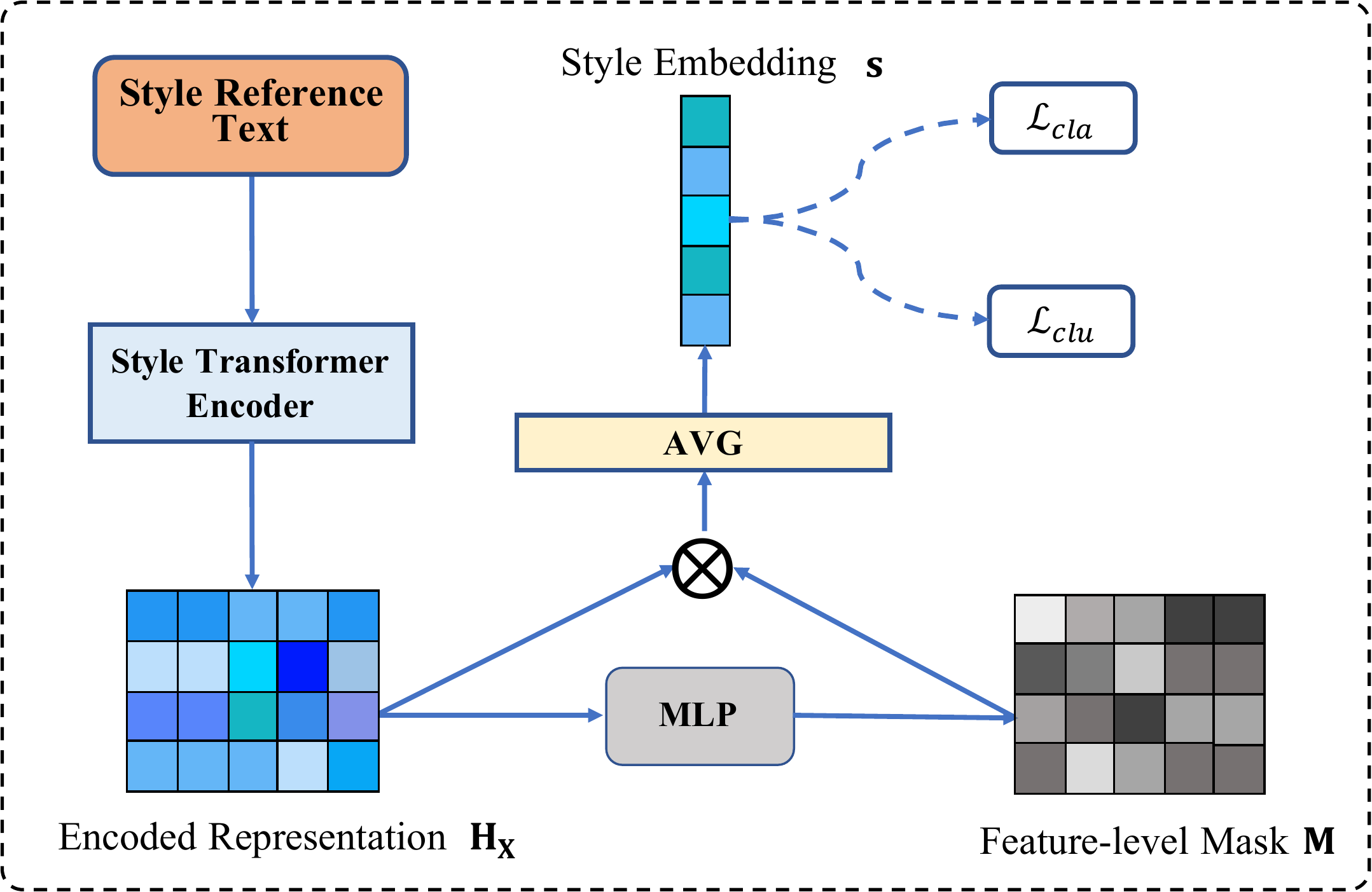}
  \caption{Illustration of the mask-based style embedding module. $\bigotimes$ denotes the elementwise multiplication operation. }
  \label{fig:style_model}
  \vspace{-0.5em}
\end{figure*}

\subsection{Mask-based Style Embedding}
Existing studies on stylized generation mainly extract the style information from the high-level sentence representation or learn the style embedding directly by omitting certain low-level unnecessary word representations. 
Nevertheless, we argue that each word in the style reference text could simultaneously contain both the content and style information. Thus, previous style extraction methods suffer from style information loss. Accordingly, we propose a derivation of the style information with an intermediate feature-level mask, as shown in Figure~\ref{fig:style_model}. In particular, we first embed the style reference text $X$ by the pretrained embedding layer of BART,
\begin{equation}
   \textbf{E}_X = BART\_Embedding(X),
\end{equation}
where $\mathbf{E}_X \in \mathbb{R}^{Q \times d_1}$ denotes the embedding of the style reference text $X$. $Q$ is the total number of tokens in text $X$, and $d_1$ is the dimension for each token embedding. Similar to the data embedding, we also employ a transformer encoder as our style encoder as follows:
\begin{equation}
    \mathbf{H}_X=Transformer_X(\textbf{E}_X),
\end{equation}
where $\mathbf{H}_X \in \mathbb{R}^{Q \times d_1}$ denotes the encoded representation of $X$. Obviously, the extracted feature $\mathbf{H}_X$ contains both content and style information, but the content is not needed by the generation model. Additionally, content may interfere with the generation model, causing content inconsistency between the generated text and input data. Therefore, we should remove the content of the input text and disentangle its style. 
We then introduce the feature-level mask $\mathbf{M} \in \mathbb{R}^{Q\times d_1}$ to derive the style embedding as follows:
\begin{equation} \label{eq11}
\left\{
 \begin{aligned} 
 &\mathbf{M} = \sigma(\mathbf{H}_X \mathbf{W}_m + \mathbf{b}_m),\\
 &\mathbf{s} = AVG(\mathbf{M} \otimes \mathbf{H}_X),
\end{aligned}
\right.
 \end{equation}
where $\sigma(\cdot)$ is the sigmoid function and $\mathbf{W}_m \in \mathbb{R}^{d_1 \times d_1}$ and $\mathbf{b}_m \in \mathbb{R}^{d_1}$ are learnable parameters. 
$\otimes$ is the elementwise product operation, and $\mathbf{s} \in \mathbb{R}^{d_1}$ is the style embedding extracted from the reference text $X$. 
To achieve accurate style embedding, we introduce two learning constraints for optimizing the parameters of the mask-based style embedding module.

\textit{Style Discrimination.} Here, we introduce the style classifier to guide style embedding learning,
    \begin{equation}
                \mathbf{\hat{g}} = softmax(\mathbf{W}_s^2(tanh(\mathbf{W}_s^1 \mathbf{s}+\mathbf{b}_s^1))+\mathbf{b}_s^2),
    \end{equation}
    where $\mathbf{W}_s^1 \in \mathbb{R}^{d_1 \times d_1}$, $\mathbf{W}_s^2 \in \mathbb{R}^ {d_1 \times N_s}$, $\mathbf{b}_s^1 \in \mathbb{R}^{d_1}$ and $\mathbf{b}_s^2 \in \mathbb{R}^{N_s}$ are trainable parameters. $\mathbf{\hat{g}}=[\hat{g}_1,\hat{g}_2,\cdots,\hat{g}_{N_s}]$ is the predicted style probability distribution. We then use the cross-entropy loss as the objective function,
    \begin{equation} \label{eq13}
    \mathcal{L}_{cla} = - \sum_{i=1}^{N_s} g_i \log \hat{g}_i.
    \end{equation}
    
\textit{Mutual Information Maximization.} 
From the information theory perspective and to encourage the learned style embedding $\textbf{s}$ to be representative, it is desirable to maximize the mutual information between the style embedding $\textbf{s}$ and the input style reference text $X$. 
Therefore, we have the following loss function according to~\cite{DBLP:conf/iclr/YuanCZHGC21}:
    \begin{align} \label{eq14}
    \mathcal{L}_{clu} =& ||\textbf{s} - \textbf{s}^u||^2 +
         \frac{1}{N_s}
         \sum_{v, v\neq u}
         {exp(-||\textbf{s} - \textbf{s}^v||^2)}, 
    \end{align}
    where we suppose that the given style reference text belongs to the $u$-th style, and $\textbf{s}^u=AVG_{j:g_j^u=1}(\mathbf{s}_j)$
    denotes the center embedding of the $u$-th style. 
 
\begin{figure*}[!t]
  \centering
  \includegraphics[scale=0.5]{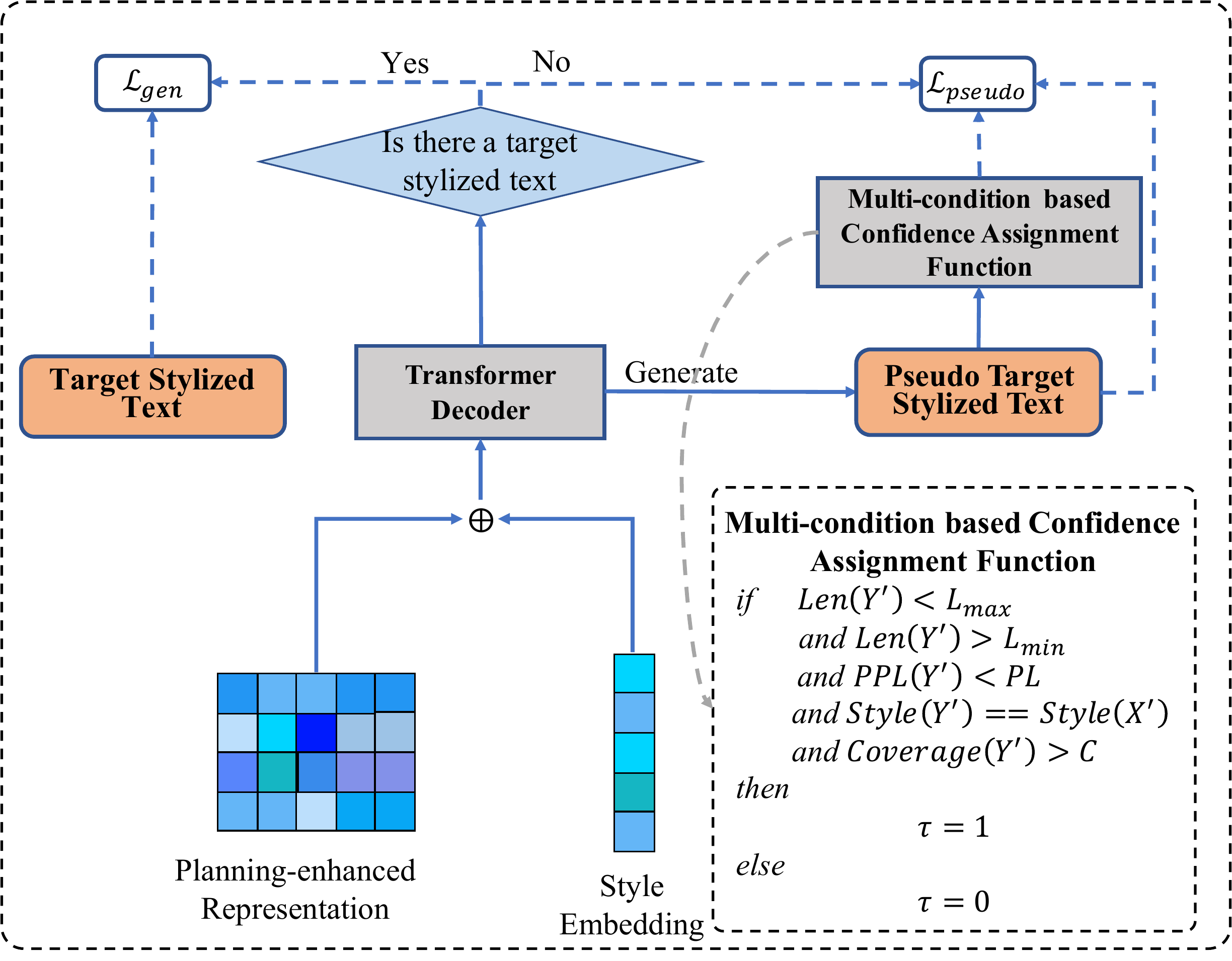}
  \caption{Illustration of the unbiased stylized text generation module. ``$\bigoplus$'' denotes the concatenation operation.}
  \label{fig:generator}
  \vspace{-0.5em}
\end{figure*}

\subsection{Unbiased Stylized Text Generation}
As shown in Figure~\ref{fig:generator}, based on the logic planning-enhanced data representation $\mathbf{H}_o$ and the style embedding $\mathbf{s}$, we generate the target text $Y$ with a transformer decoder as follows:
\begin{equation}
    \hat{\textbf{y}}_t=Transformer_g([\mathbf{H}_o; \mathbf{s}], \mathbf{E}_{Y_{<t}}),
    \label{generator}
\end{equation}
where $Y_{<t}=\{Y_1,\cdots, Y_{t-1}\}$ refers to the previous $t-1$ tokens, and $\mathbf{E}_{Y_{<t}}$ denotes the embedding of the previous tokens derived by Eqn. ($\ref{embedding}$). $\hat{\textbf{y}}_t \in \mathbb{R}^{d_1}$ is the output hidden state vector encoding the style, planning, and previously generated token information utilized for predicting the $t$-th token.

For optimization, we introduce the following loss function for stylized data-to-text generation:
\begin{equation} \label{prob}
 \left\{
 \begin{aligned}
  &\mathcal{L}_{gen} = -\log{P(Y|\mathcal{P}, X)}=-log{\prod \limits_{t} P(Y_t|Y_{<t}, \mathbf{H}_o, \mathbf{s})},\\
  &P(Y_t|Y_{<t}, \mathbf{H}_o, \mathbf{s})=softmax_{Y_t}(\textbf{W}_Y \hat{\textbf{y}}_t +\textbf{b}_Y),
  \end{aligned}
 \right.
 \end{equation}
where $Y_t$ is the $t$-th generated token of the target text, while $\textbf{W}_Y \in \mathbb{R}^{|\mathcal{V}| \times d_1}$ and $\textbf{b}_Y \in \mathbb{R}^{|\mathcal{V}|}$ are parameters to be learned. $\mathcal{V}$ is the token vocabulary, and $|\mathcal{V}|$ denotes its size. 
$P(Y_t|Y_{<t}, \mathbf{H}_o, \mathbf{s})$ represents the probability that the $t$-th generated token is the ground-truth token $Y_t$.


\textit{Pseudo Triplet Augmentation.}
In fact, given data, it is more likely that we only have text corresponding to a single style. For example, given a product, we may only have its formal text and not its informal text. Therefore, optimizing the model with these types of biased training samples may lead to biased text generation, \textit{i.e.}, a stylized text generator that relies only on the content of the given data to generate the target text and overlooks the specified style information. In other words, once our generator encounters the same testing data, it tends to generate the text of a specific style according to the training data, regardless of the given style reference text. 
To address this issue, for each data, we augment a pseudo-stylized text for its missing style to promote the diversity of the training dataset. Suppose that we have the target stylized text $Y$ of style $s$ for the given data $\mathcal{P}$ but no stylized text $Y'$ of style $s'$. We then select a style reference text $X'$ of the style $s'$ and feed it together with the data $\mathcal{P}$ to our transformer decoder in Eqn. ($\ref{generator}$) to obtain a target text candidate $Y'$ according to Eqn. ($\ref{prob}$).

\textcolor{black}{Actually, the augmented texts with the model may be of low quality. Therefore, blindly adding these augmented data will bring noise and degrade the performance of the model. Firstly, the style of augmented text should be accurate. Second, the content consistency should be high. Next, the augmented texts should be fluent. Last but not least, the length of the augmented text should in the distribution of the origin training set. Based on these requirements, we choose style, coverage, perplexity, and text length as four metrics to test whether the augmented data is high quality. To be specific, } 
we define a multi-condition-based confidence assignment function as follows:
\begin{align} \label{confi}
    \tau = & \mathbbm{1}(Len(Y') < L_{max} \; and \; Len(Y') > L_{min}
     \; and \; PPL(Y') < PL \; and \; Style(Y') == Style(X') \; \notag \\ & 
    and \; Coverage(Y', \mathcal{P}) > C),
\end{align}
where $\tau\in \{0,1\}$ is the binary confidence for the generated text $Y'$, which depends on the length, perplexity~(\textit{i.e.}, PPL)~\cite{chen1998evaluation}, style and coverage~\cite{DBLP:conf/emnlp/ShaoHWXZ19} of the generated text $Y'$.  $L_{max}$, $L_{min}$, $PL$, and $C$ are the predefined thresholds for the corresponding conditions.

Once the generated text meets all the conditions in Eqn. ($\ref{confi}$), we deem the triplet $(\mathcal{P}, X', Y')$ as a reliable pseudo triplet. Accordingly, we introduce the additional loss function for the pseudo triplet as follows:
\begin{align} \label{pseudo}
    \mathcal{L}_{pseudo} = - \tau \log {P(Y'| \mathcal{P}, X')}.
\end{align}



\begin{algorithm}[!t]
\caption{Training Process}
\label{alg:training}
\begin{flushleft}
\hspace*{0.02in} {\bf Input:}
Training triplets $\mathcal{D}$, model $\mathcal{M}$, the batch size $B$, the number of epochs $N_{epoch}$, the number of iterations $N_{iter}$, hyperparameters $L_{max}$, $L_{min}$, $PL$, $C$, $\alpha$, $\beta$, $\gamma$, $\delta$, the number of styles $N_s$.\\
\hspace*{0.02in} {\bf Require:} The parameters to be learned $\boldsymbol{\Theta}$.
\end{flushleft}
\begin{algorithmic}[1]
\STATE {Initialize parameters $\boldsymbol{\Theta}$.}
\FOR{$e=1$ to $N_{epoch}$}
\FOR{$t=1$ to $N_{iter}$}
\FOR{$i=1$ to $N_s$}
\STATE Randomly sample $B$ examples from $\mathcal{D}$ for the $i$-th style.
\ENDFOR
\STATE Compute the loss function of planning according to Eqn. ($\ref{eq7}$).
\STATE Compute $\mathcal{L}_{cla}$ and $\mathcal{L}_{clu}$ according to  Eqn. ($\ref{eq13}$) and  Eqn. ($\ref{eq14}$), respectively.
\STATE Compute the generation loss function according to Eqn. ($\ref{prob}$).
\FOR{each training triplet $(\mathcal{P}, X, Y)$ in the batch}
\STATE Generate a pseudo triplet $(\mathcal{P}, X', Y')$
\STATE Compute confidence $\tau$ for the generated pseudo triplet according to Eqn. ($\ref{confi}$).
\ENDFOR
\STATE Compute the loss function for pseudo triplets according to Eqn. ($\ref{pseudo}$).
\STATE Update the parameters $\boldsymbol{\Theta}$ of the model $\mathcal{M}$ by Eqn. ($\ref{allloss}$).
\ENDFOR
\ENDFOR
\RETURN The learned model $\mathcal{M}$.
\end{algorithmic}
\end{algorithm}

\subsection{Training and Inference}
For training, we combine all the loss functions as follows:
\begin{equation}
\label{allloss}
    \mathcal{L} =   \mathcal{L}_{gen} + \alpha \mathcal{L}_{plan} + \beta \mathcal{L}_{cla} + \gamma \mathcal{L}_{clu} + \delta \mathcal{L}_{pseudo},
\end{equation}
where $\alpha$, $\beta$, $\gamma$, and $\delta$ are hyperparameters used for balancing the effect of each component on the entire training process. The overall optimization procedure is briefly summarized in Algorithm~\ref{alg:training}.
During the inference phase, given a testing sample, we first employ the well-trained logic planner to predict its attribute organization plan as $\hat{L}$. Then, according to $\hat{L}$, we can derive the planning-enhanced data representation $\mathbf{\hat{H}}_o$ by Eqn. ($\ref{planning}$). Finally, we adopt the greedy search algorithm to generate the target text $\hat{Y}$ according to Eqn. ($\ref{prob}$). 


\section{Experiments} \label{sec:exp}
To justify our model, we conduct extensive experiments on a real-world dataset collected from the largest Chinese e-commerce platform Taobao and answer the following research questions:\\
\textbf{RQ1.} Can existing methods be used to solve the proposed task?\\
\textbf{RQ2.} How do the components of the StyleD2T model affect its performance?\\
\textbf{RQ3.} What is the qualitative performance of the StyleD2T model?

\subsection{Dataset}
To verify the effectiveness of StyleD2T, we collected a Chinese stylized data-to-text dataset named TaoStyle from TaoBao, which contains $20,000$ formal advertising texts of products written by qualified experts on the Weitao platform and $11,728$ informal advertising transcripts exacted from live broadcast video streams on TaoBao using automatic speech recognition (ASR)\footnote{\url{https://help.aliyun.com/document_detail/90727.html}.}. 
\textcolor{black}{In fact, some noisy transcripts may be introduced by ASR in the dataset construction process. Thus we conducted manual corrections for the noisy transcripts. The to-be-corrected problems include broken sentences, word errors, and word repetition. Meanwhile, during the data collection process, we ignored the text whose length is less than $10$ or more than $300$.}
Notably, each stylized text is linked to a product. Accordingly, for each stylized text, we can derive a set of attribute-value pairs as the nonlinguistic input data by using the corresponding product's attribute properties from its webpage. 
For keeping the consistency between the input data and the target text, we only retain the attribute-value pairs from the product's webpage that appear in the target stylized text. 
Then, for each data and its corresponding target text, we randomly sample stylized text which has the same style with the target text as the style reference to make the (data, style reference text, target text) triplets.

\begin{figure} \centering 
\subfigure[Informal texts.] {
 \label{fig:a_cloud}     
\includegraphics[width=0.45 \columnwidth]{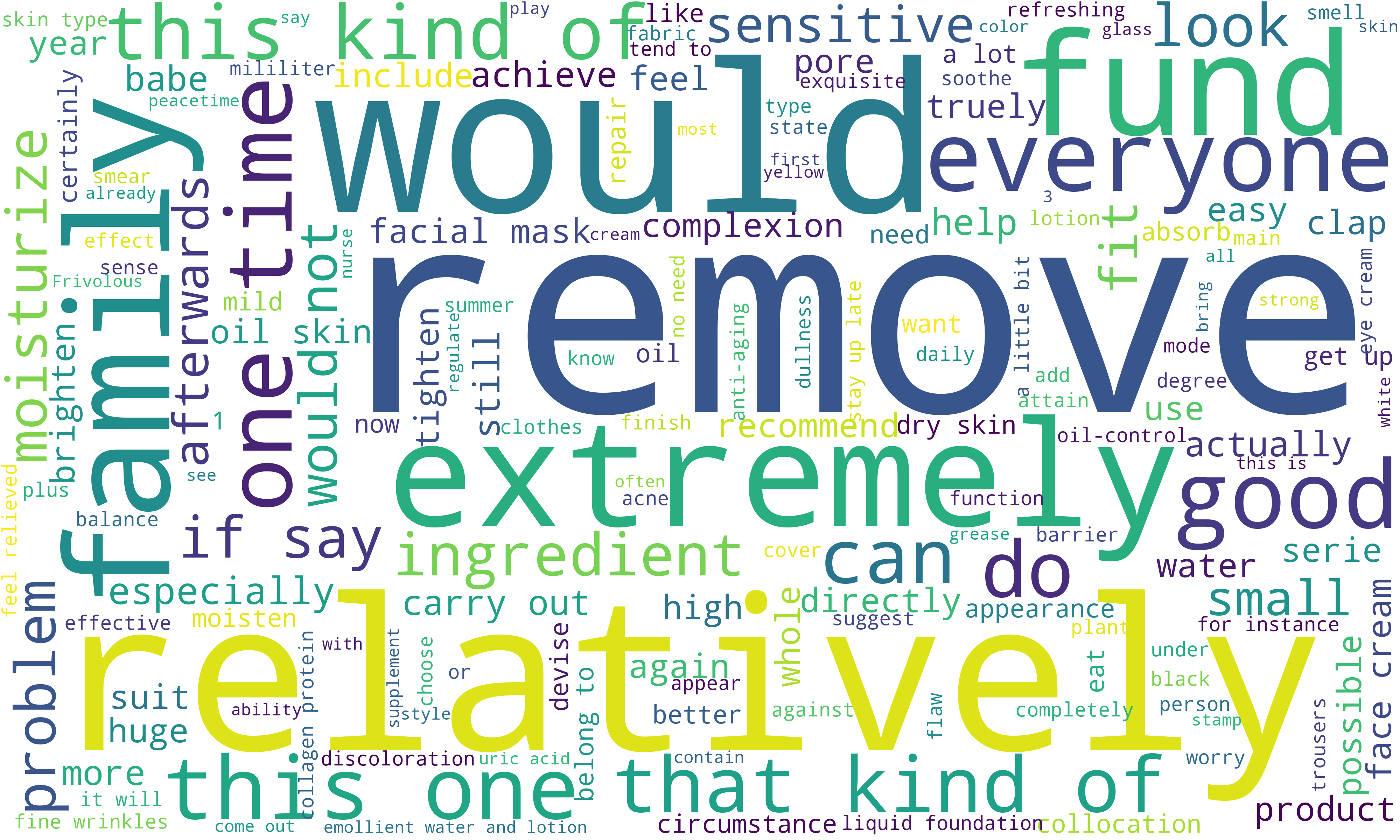}  
}     
\subfigure[Formal texts.] { 
\label{fig:b_cloud}     
\includegraphics[width=0.45 \columnwidth]{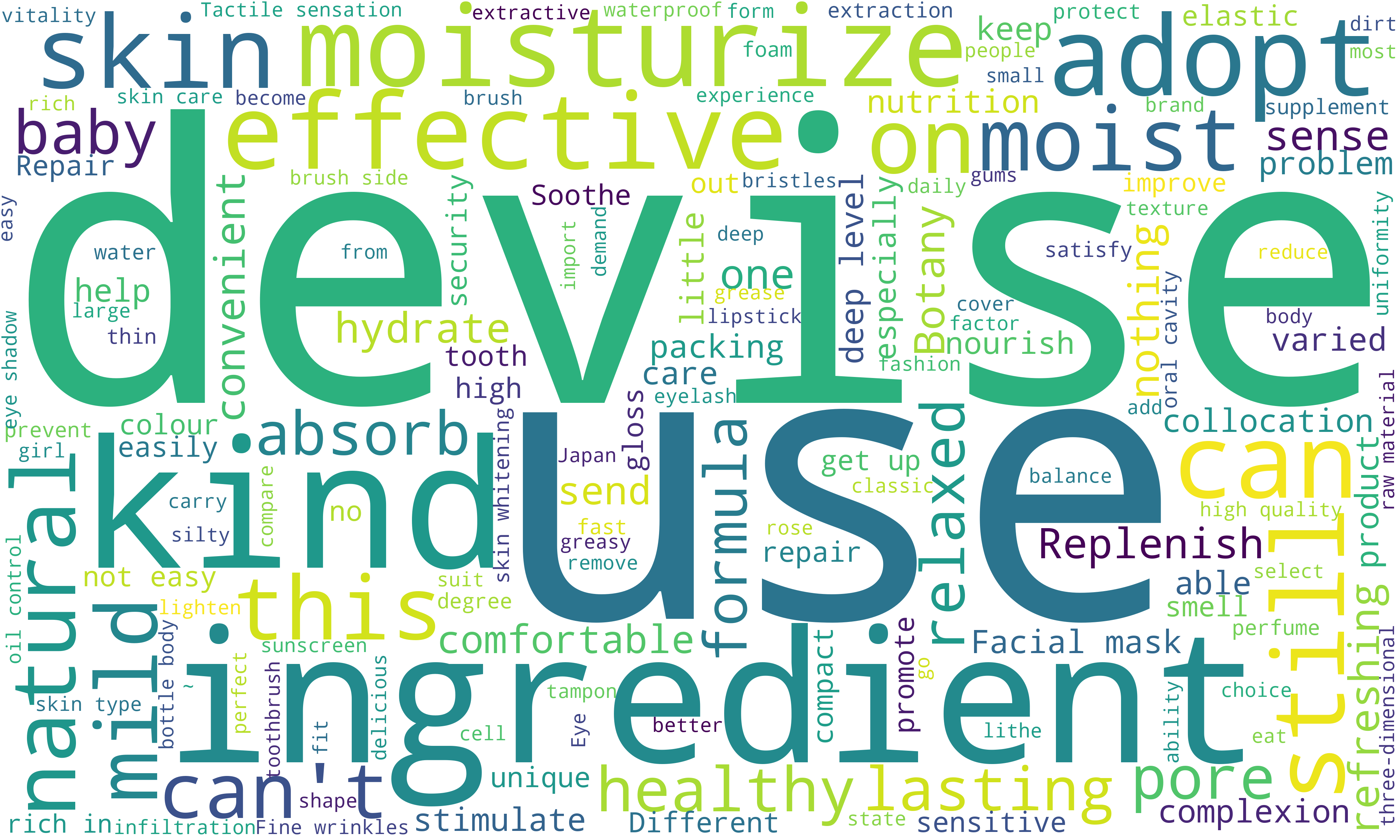}     
}    
\caption{ Word clouds of two styles of texts. }     
\vspace{-0.5em}
\label{fig:cloud}     
\end{figure}
\begin{figure} \centering 
\subfigure[Informal texts.] {
 \label{fig:a_len}     
\includegraphics[width=0.48\columnwidth]{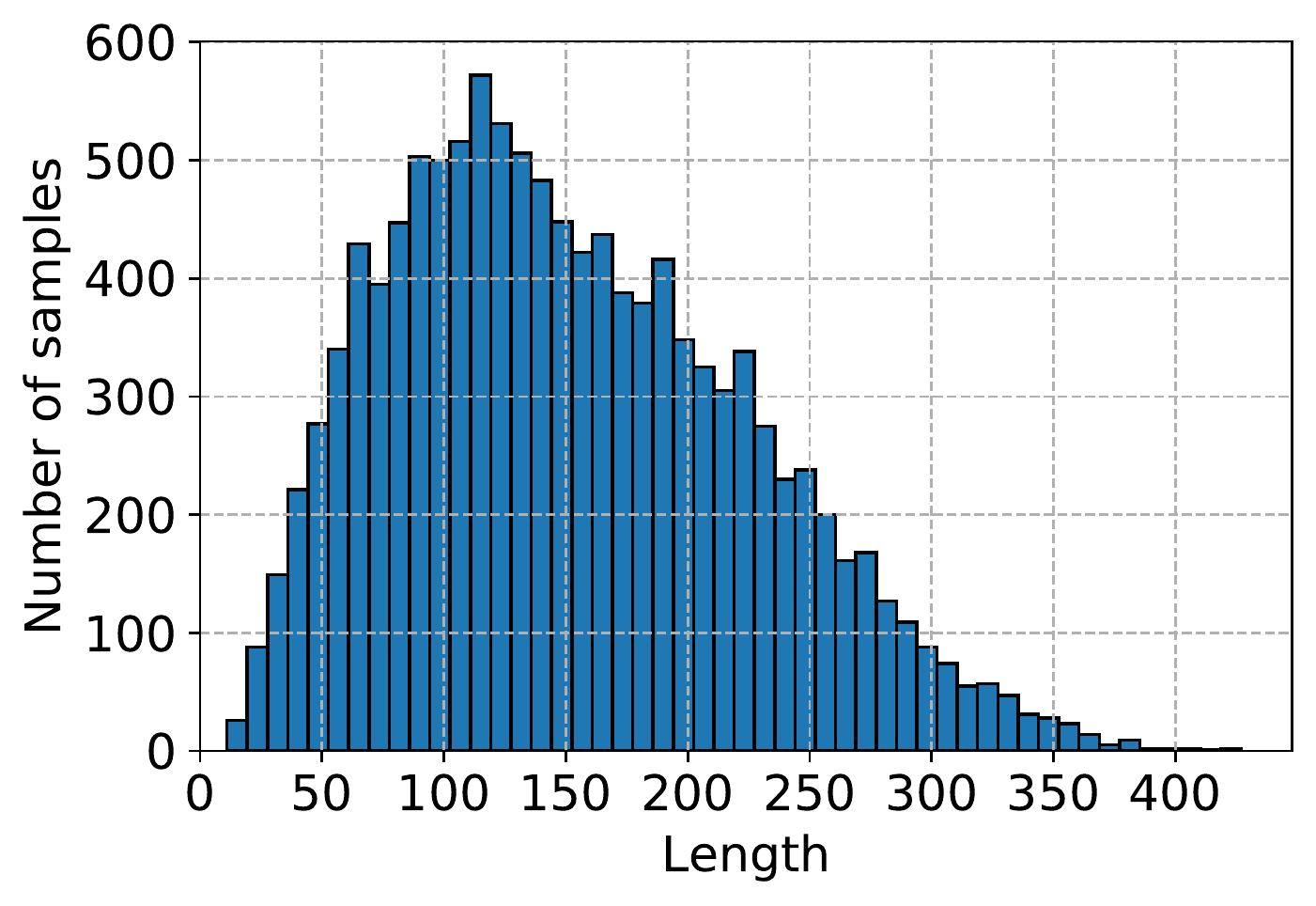}  
}     
\subfigure[Formal texts.] { 
\label{fig:b_len}     
\includegraphics[width=0.48 \columnwidth]{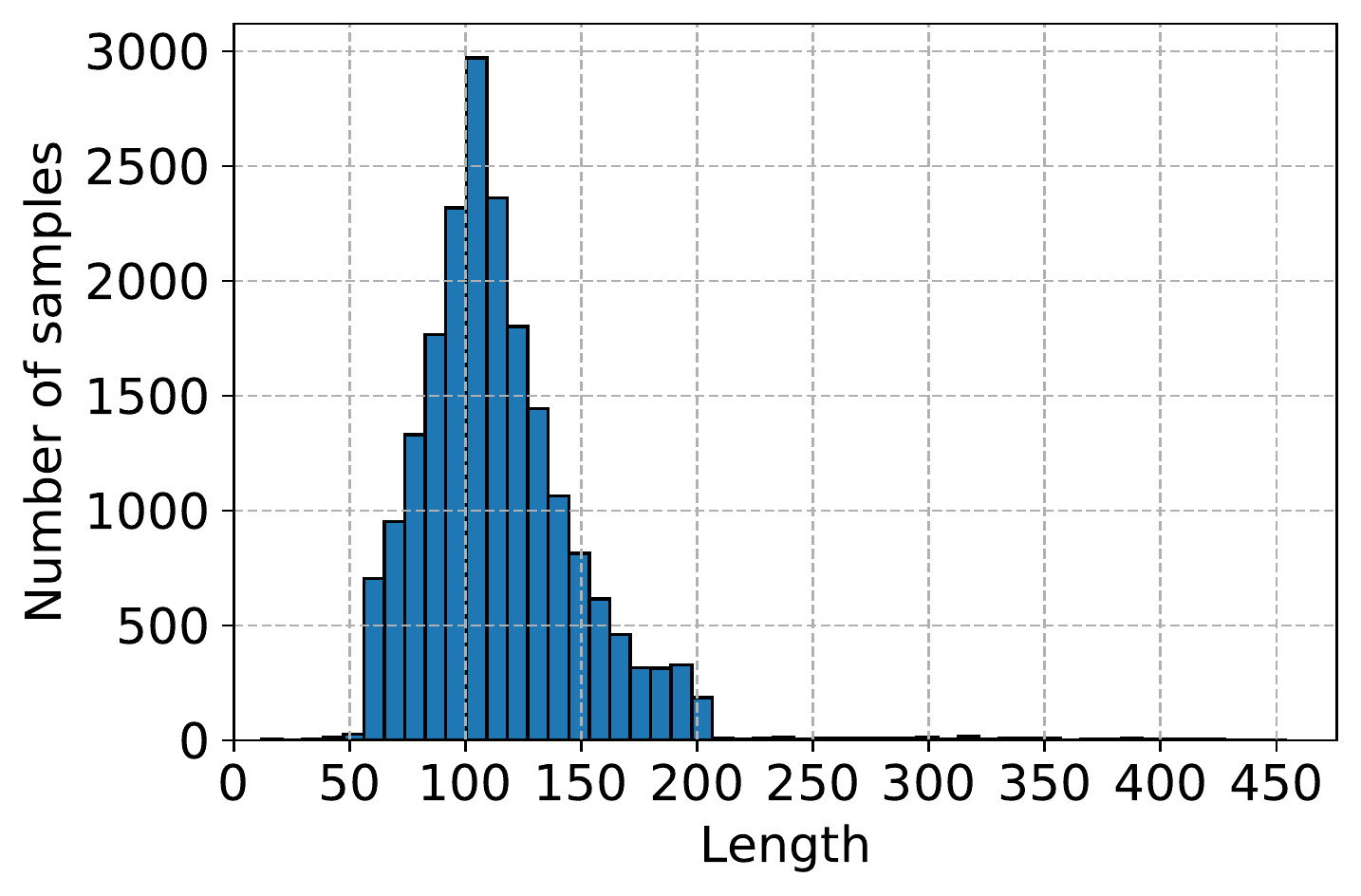}     
}    
  \vspace{-1em}
\caption{The distributions of text length for two styles of texts. }     
\label{fig:len}     
  \vspace{-0.5 em}
\end{figure}

Ultimately, \textcolor{black}{we randomly split our corpus and derive a training set, a validation set and a test set, comprising $27,728$, $2,000$, and $2,000$ triplets, respectively.} Table~2 summarizes the statistics of our dataset. In addition, the average token count of the target text and the average number of input attribute-value pairs are $115$ and $8.1$ for the informal style and $151$ and $4.6$ for the formal style, respectively. The total numbers of attributes and values are $108$ and $26,915$ for the formal style and $210$ and $9,567$ for the informal style, respectively.
To intuitively obtain the word distribution difference between informal and formal texts, we show the word clouds over two styles of texts in Figure~\ref{fig:cloud}. As seen, the words ``remove'', ``would'', and ``relatively'' most frequently appear in informal texts, while ``devise'', ``use'' and ``ingredient'' often appear in formal texts.
In addition, we also visualize the text length distributions of the two styles of texts in Figure~\ref{fig:len}. Compared with informal texts, the text length distribution of formal texts is more concentrated. This suggests that the informal texts are looser, as compared with formal texts.

\begin{table}[]
\setlength{\abovecaptionskip}{0cm}
\setlength{\belowcaptionskip}{0cm}
\caption{The statistics of TaoStyle dataset. \#Total Attr/\#Total Val: the total number of attributes / attribute values. \#Avg Attr-val: the average number of input attribute-value pairs. \#Avg Len: the average length of target stylized text. }
\label{statistics}
\centering
\setlength{\tabcolsep}{5pt}{
\begin{tabular}{|c|c|c|c|c|c|c|c|}
\hline
\textbf{Style} & \textbf{ \ \ Train \ } & \textbf{\ \ Valid\ } & \textbf{\ \ Test  \ } &\textbf{\#Total Attr} & \textbf{\#Total Val} &\textbf{\#Avg Attr-val} & \textbf{\#Avg Len}    \\ \hline
Informal & 9,728 & 1,000 & 1,000 & 108& 26,916 & 8.1 & 115 \\ \hline
Formal &  18,000& 1,000 & 1,000 & 210& 9,567 & 4.6 & 151 \\ \hline
\end{tabular}
}
\vspace{-1em}
\end{table}

\subsection{Evaluation Metrics.} 
For the comprehensive evaluation, we conducted both quantitative and qualitative evaluations. 
\subsubsection{Quantitative Evaluation}
To quantitatively evaluate the quality of generated texts, we used the following four widely used metrics: (1) \textbf{Style Accuracy} refers to the style accuracy of the generated text\footnote{Similar to~\cite{DBLP:conf/acl/DaiLQH19}, we train a style classifier with the training set using fastText~\cite{DBLP:conf/eacl/GraveMJB17}.}, 
(2) \textbf{Coverage}~\cite{DBLP:conf/emnlp/ShaoHWXZ19} measures the average proportion of input attribute-value pairs covered by the generated text, (3) \textbf{\mbox{ROUGE-L}}~\cite{lin2004rouge} evaluates the similarity between the generated text and the target text based on their longest common subsequences, and (4) \textbf{BLEU-4}~\cite{DBLP:conf/acl/PapineniRWZ02} measures the difference between the generated text and the ground truth based on the $4$-gram. 

For the unbiased stylized data-to-text generation, although we do not simultaneously have both the ground-truth formal and informal texts for each testing data, we still generate a formal text and an informal text for each input data. Then, we compute BLEU-4 and ROUGE-L, which require ground-truth texts based on both the generated and ground-truth texts, while the other metrics are derived only by the generated texts.

\subsubsection{Qualitative Evaluation}
Apart from the quantitative evaluation, we also consider the qualitative evaluation, as it has been reported~\cite{DBLP:conf/eacl/Schluter17,DBLP:conf/aaai/FuTPZY18,DBLP:conf/naacl/MirFOR19} that the quantitative evaluation is unreliable in generation tasks. 
In this part, we first select $100$ testing samples for evaluation. Then, we employ $40$ volunteers to manually evaluate the quality of the generated texts of different models for the testing samples. Specifically, for each testing sample, we generate both formal and informal texts, and we have $200$ cases for evaluation. Every case is annotated $5$ times. These cases are randomly distributed to the $40$ volunteers. Namely, each volunteer is asked to evaluate $25$ cases. 
For each case, the volunteer needs to score all the generated texts of different models according to the following three dimensions: (1) \textbf{Style Consistency} is used for judging whether the style of the generated text meets the given requirements, (2) \textbf{Content Consistency} is used for evaluating the consistency between the input data and generated text, and (3) \textbf{Text Fluency} is used for assessing the readability of generated texts. 
The score range for each metric is $\{1, 2, 3, 4, 5\}$, where $1$ refers to ``very bad'' and $5$ corresponds to ``very good''. Notably, volunteers, who were all native Chinese speakers, did not know which model was used for generating the text. 



\subsection{Implementation Details}
We train our model on a Tesla V100 GPU, and the batch size is set to 8 for each style. We use Adam as the optimizer, and the learning rate is set to $1e$$-$$4$.
Similar to BART, the number of layers of all transformer-based models (\textit{i.e.}, the style transformer encoder, data transformer encoder and transformer decoder) is set to $6$. Their hidden size, embedding size, and positional embedding size are all set to $768$, while the number of attention heads is set to $12$. $d_1$ and $d_2$ are set to $768$ and $256$. Our embedding layer (\textit{i.e.}, $BART\_Embedding(\cdot)$) is also trainable. $|\mathcal{V}|$ is $21,128$. $M$ is set to $1$. Since our dataset has texts of two styles, $N_s$ is set to $2$. 
Hyperparameters $\alpha$, $\beta$, $\gamma$, and $\delta$ are set to $1$, $1$, $1$, and $0.1$, respectively. \textcolor{black}{We used the grid search strategy to find the optimal hyperparameters. Specifically, hyperparameters $\alpha$, $\beta$, $\gamma$, and $\delta$ are searched within [0, 1] with a step size of 0.1. Because ninety-nine percent of the length of texts in our dataset is in the range [60, 160], $L_{min}$ and $L_{max}$ are set to $60$ and $160$.} $PL$ and $C$ in the multi-condition-based confidence assignment function are set to $50$ and $0.95$, respectively. 
To obtain the perplexity and style of the generated text in Eqn. ($\ref{confi}$), similar to the text style transfer studies~\cite{DBLP:conf/acl/DaiLQH19,DBLP:conf/naacl/LiJHL18}, the $5$-gram language model is trained on our training dataset by KenLM\footnote{\url{https://kheafield.com/code/kenlm/}.} and the style classifier is trained on our training dataset with fastText\footnote{\url{https://fasttext.cc/docs/en/supervised-tutorial.html}.} to implement $PPL(\cdot)$ and $Style(\cdot)$.


\subsection{On Model Comparison (RQ1)}

Since there is no existing method for the stylized data-to-text task, we stack existing general data-to-text generation methods and text style transfer methods to derive our baselines. The data-to-text method can generate a base text for the given data, while the text style transfer model can further render the generated base text into the desired style. 
In particular, we adopt four data-to-text methods as follows:

\begin{table}[]
\setlength{\abovecaptionskip}{0cm}
\setlength{\belowcaptionskip}{0cm}
\caption{Quantitative evaluation performance comparison with baselines. The bold font indicates the best result and the second best result is underlined. \textcolor{black}{ * denotes that the $p$-value of the significant test between our result and the best baseline result is less than 0.01.} ``Improvement. $\uparrow$'' refers to  the relative improvement by our model over the best baseline result. }
\label{RQ1}
\centering
\setlength{\tabcolsep}{1mm}{
\begin{tabular}{l|ccccc}
\hline
\textbf{Model} & \textbf{Style Accuracy} & \textbf{Coverage} & \textbf{ROUGE-L}&  \textbf{BLEU-4}      \\ \hline \hline
{Seq2seq+StyleT}&  68.90& 72.84& 17.44 & 0.30      \\
{Seq2seq+NAST}& 87.35& 68.00 & 17.44&  0.30       \\
{BART+StyleT}& 71.03 & 88.07 & 25.38 & 5.28  \\ 
{BART+NAST} & \underline{89.48} & 88.07  & 25.38 & 5.28   \\ 
{PHVM+StyleT}& 50.08& 58.49 & 17.67 & 1.05    \\
{PHVM+NAST} & 68.52 & 38.42 & 17.67 & 1.05   \\
{PlanGen+StyleT} & 67.18 & \underline{88.40}& \underline{25.35} & \underline{5.31}   \\
{PlanGen+NAST} & 85.62 &  68.25 & \underline{25.35} & \underline{5.31}     \\ \hline
{StyleD2T} & \textbf{97.49}* &\textbf{93.44}* & \textbf{26.45}* & \textbf{5.77}* \\
{Improvement. $\uparrow$}  & 8.95\% & 5.70\%& 4.34\% & 8.66\%   \\
\hline
\end{tabular}
}
\vspace{-0.5em}
\end{table}

\begin{table}[]
\setlength{\abovecaptionskip}{0cm}
\setlength{\belowcaptionskip}{0cm}
\caption{The average scores of the qualitative evaluation results for three metrics. The best results are in bold. \textcolor{black}{ * denotes that the $p$-value of the significant test between our model result and the best baseline result is less than 0.01.}}
\label{table:Human}
\centering
\setlength{\tabcolsep}{1mm}{
\begin{tabular}{l|ccc}
\hline
\textbf{Model} & \textbf{Style Consistency} & \textbf{Content Consistency}  & \textbf{Text Fluency}  \\ \hline \hline
{BART+StyleT} & 3.50 & 3.80 &3.22   \\ 
{BART+NAST} & 3.50 & 4.18 & 3.89  \\ 

{PlanGen+StyleT} & 3.43 & 3.71 &  3.20  \\
{PlanGen+NAST} & 3.38  & 4.12 &4.10   \\ \hline
{StyleD2T} & \textbf{4.21}* & \textbf{4.31}* & \textbf{4.28}* \\
\hline
\end{tabular}
}
\end{table}
(1) \textbf{Seq2Seq}~\cite{DBLP:conf/emnlp/RushCW15}, an LSTM-based model with an attention mechanism. This method utilizes a local attention-based model to generate a target text of each word from the input text.

(2) \textbf{BART}~\cite{DBLP:conf/acl/LewisLGGMLSZ20}, a denoising autoencoder for pretraining sequence-to-sequence models. It is trained by reconstructing the original text of corrupted text with an arbitrary noising function. It has achieved state-of-the-art performance in many natural language generation tasks.

(3) \textbf{PHVM}~\cite{DBLP:conf/emnlp/ShaoHWXZ19}, a planning-based hierarchical variational model for data-to-text generation. It first selects the input items contained in a sentence and then generates the corresponding sentence.

(4) \textbf{PlanGen}~\cite{DBLP:conf/emnlp/SuVWFC21}, a plan-then-generate framework to improve the controllability of data-to-text models. It is a multistage framework and utilizes the slot keys from a table as a planning signal. 

In addition, we used two text style transfer methods:

(a) \textbf{StyleT}~\cite{DBLP:conf/acl/DaiLQH19}, a transformer-based style transfer model for unpaired text. This is the first work to apply the transformer architecture to style transfer tasks. 

(b) \textbf{NAST}~\cite{DBLP:conf/acl/HuangCWGZH21}, a nonautoregressive generator for unsupervised text style transfer tasks. It mainly focuses on the connections between aligned words and learns the word-level transfer between styles.

Finally, based on these methods, we derive the following eight baselines: (1) \textbf{Seq2seq+StyleT}, (2) \textbf{Seq2seq+NAST}, (3) \textbf{BART+StyleT}, (4) \textbf{BART+NAST}, (5) \textbf{PHVM+StyleT}, (6) \textbf{PHVM+NAST}, (7) \textbf{PlanGen+StyleT}, and (8) \textbf{PlanGen+NAST}. 

\textbf{Quantitative Evaluation}. Table~\ref{RQ1} shows the performance comparison between our model and the baselines. As shown by this table, our StyleD2T model exceeds all baselines on all metrics. In particular, coverage is significantly improved (an increase of 5.70\%) when using our StyleD2T model compared with the best baseline, which suggests the superiority of our model in maintaining the semantic consistency of the generated text for the given data. Meanwhile, the style accuracy is improved from 89.48\% using the best baseline model to 97.49\% using our model, which indicates that StyleD2T achieves superior performance in the context of biased training samples. In addition, our StyleD2T also achieves improvements on BLEU and ROUGE-L due to the three modules that we designed. These observations confirm the superiority of our model over the existing methods.

\begin{figure} \centering 
 \subfigure[Style Consistency.] { 
\label{fig:style}     
\includegraphics[width=0.31 \columnwidth]{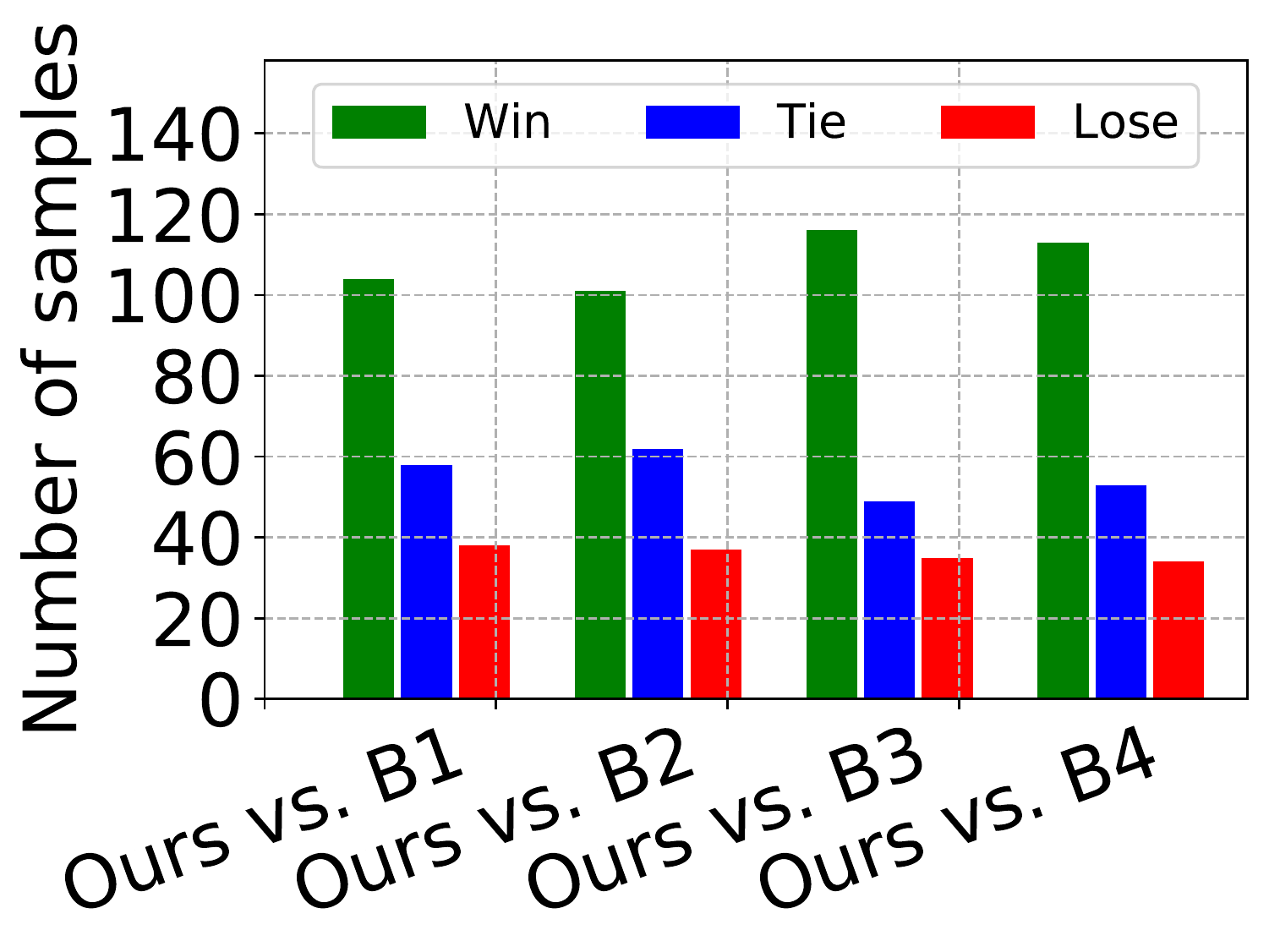}     
}    
\subfigure[Content Consistency.] { 
\label{fig:content}     
\includegraphics[width=0.31 \columnwidth]{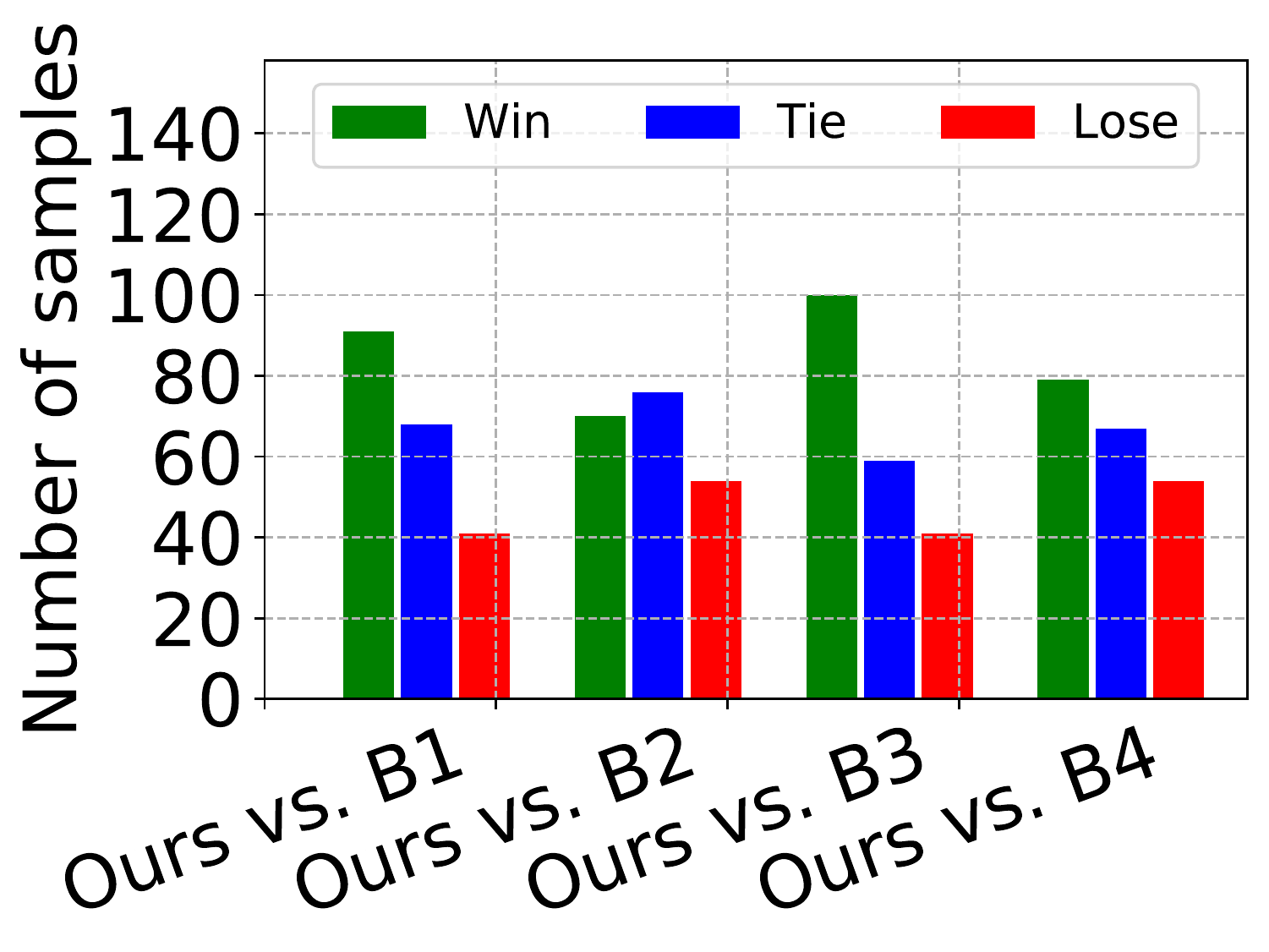}     
}    
\subfigure[Text Fluency.] {
 \label{fig:fluency}     
\includegraphics[width=0.31 \columnwidth]{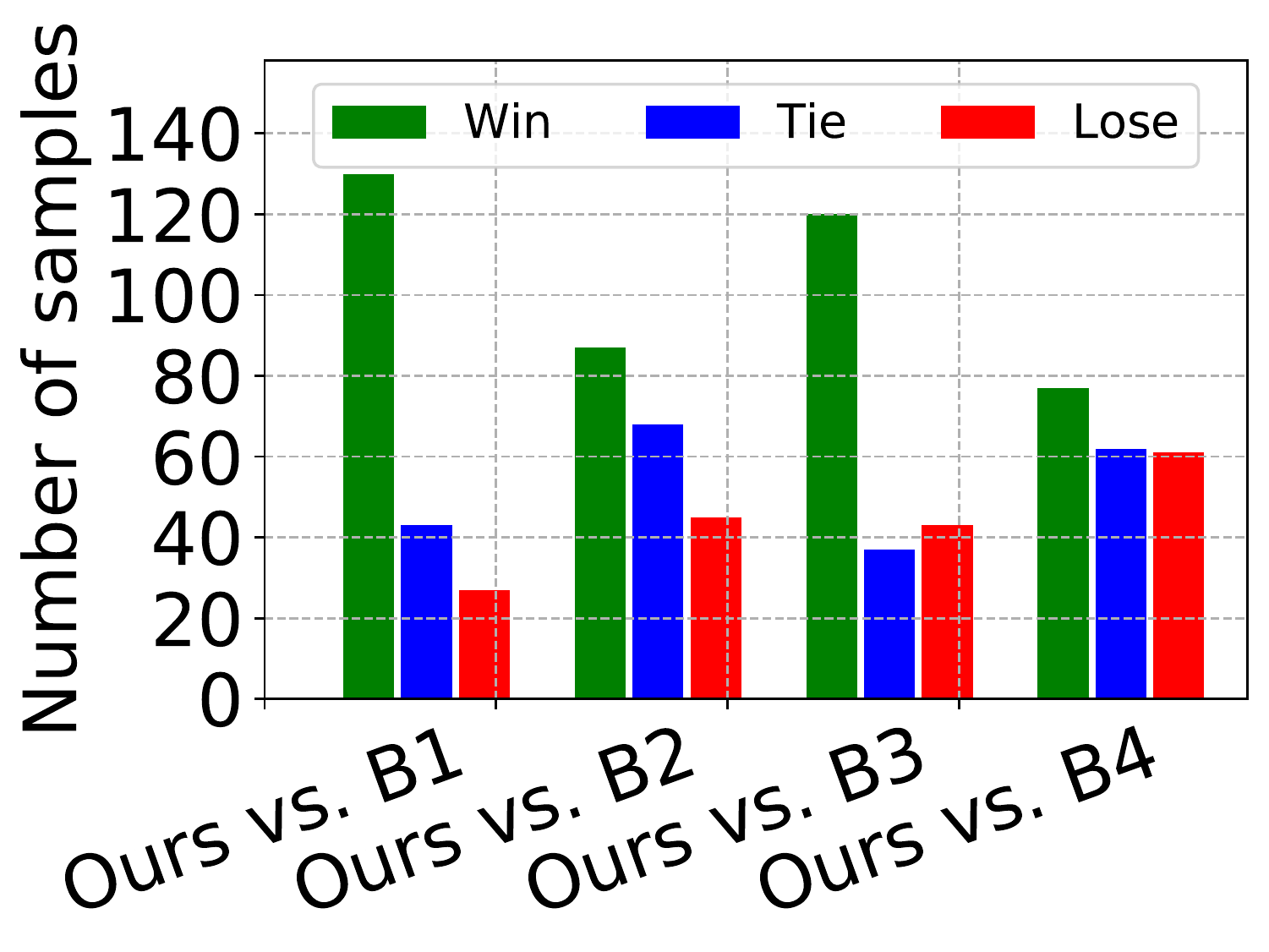}  
}     
  \vspace{-1em}
\caption{Comparison results between StyleD2T and the baseline models for three qualitative evaluation metrics. ``B1'', ``B2'', ``B3'', and ``B4'' denote the baseline BART+StyleT, BART+NAST, Plangen+StyleT, and Plangen+NAST, respectively. ``Win'', ``Tie'', and ``Lose'' indicate whether the score of StyleD2T is higher than, equal to, and lower than the baseline models, respectively. }       \vspace{-1em}

\label{fig:human}     
\end{figure}
\textbf{Qualitative Evaluation}. For the qualitative evaluation, we choose StyleD2T and the four best baselines (\ie BART+StyleT, BART+NAST, Plangen+StyleT, and Plangen+NAST). Table~\ref{table:Human} shows the average scores of Style Consistency, Content Consistency, and Text Fluency for all testing samples of each model. We observed that our StyleD2T model outperforms the baselines in all three dimensions (\ie Style Consistency, Content Consistency and Text Fluency), which is consistent with the aforementioned quantitative evaluation results. 
This further indicates that our StyleD2T model has a more powerful stylized generation ability than the baselines. It is worth noting that our model achieves more improvement on the Style Consistency metric over the baselines compared with the other two metrics (\textbf{i.e.}, Content Consistency and Text Fluency). This suggests that our model is more robust towards unbiased stylized text generation and the necessity of introducing pseudo target texts.

To gain a more intuitive understanding of the qualitative evaluation results, we also report the comparison results between our StyleD2T model and the four baseline models in Figure~\ref{fig:human}. In particular, for each pair of model comparisons and each metric, we show the number of samples where our model achieves higher (denoted as ``Win''), equal (denoted as ``Tie''), and lower scores (denoted as ``Lose'') compared with the baselines.
As seen, StyleD2T outperforms all baselines across different evaluation metrics, as the number of ``Win'' cases is always significantly larger than that of ``Lose'' cases in each pair of model comparisons. In addition, on average, the number of ``Win'' cases is the largest for the Style Consistency metric compared with the other two metrics. This is consistent with the results in Table~\ref{table:Human}.

\textcolor{black}{\textbf{Efficiency Comparison.} To learn the efficiency of our model, we compared the inference speed of our model and baselines in Table~\ref{efficiency}. 
Notably, our model significantly surpasses the most efficient two baselines in terms of the style accuracy and coverage, as shown in Table~\ref{RQ1}. 
}

\begin{table}[]
\setlength{\abovecaptionskip}{0cm}
\setlength{\belowcaptionskip}{0cm}
\caption{\textcolor{black}{Efficiency comparison with baselines. \textbf{Time} is the average generation time consumption of samples in the testing set.}}
\label{efficiency}
\centering
\setlength{\tabcolsep}{1mm}{
\begin{tabular}{c|c|c|c|c|c}
\hline
\textbf{Model} & \textbf{Time} & \textbf{Model} & \textbf{Time} & \textbf{Model} & \textbf{Time}      \\ \hline \hline
{Seq2seq+StyleT}& 2.6s & {BART+NAST} & 0.4s & {PlanGen+StyleT} & 1.2s     \\
{Seq2seq+NAST}& 1.9s  &  {PHVM+StyleT}& 2.1s & {PlanGen+NAST} & 0.5s  \\
{BART+StyleT}& 1.6s  & {PHVM+NAST} & 1.5s & {StyleD2T} & 0.6s \\ \hline
\end{tabular}
}
\vspace{-0.5em}
\end{table}
\begin{table}[]
\setlength{\abovecaptionskip}{0cm}
\setlength{\belowcaptionskip}{0cm}
\caption{Ablation study results. The best results are in bold. \textcolor{black}{ * denotes that the $p$-value of the significant test between our model result and the best  model's variant result is less than 0.01.}} 
\label{RQ4}
\centering
\setlength{\tabcolsep}{1mm}{
\begin{tabular}{l|ccccc}
\hline

\textbf{Model}  & \textbf{Style Accuracy} & \textbf{Coverage}   & \textbf{ROUGE-L} & \textbf{BLEU-4} \\ \hline \hline
{w/o-Style} & 89.67 &  90.25 & 25.50  & 5.25 \\
{w/o-StyleCon}& 89.83 &92.27  & 25.44 & 5.10\\
{w/o-Planner} & 93.23 & 86.95 & 25.83  & 5.16 \\
{w/o-Graph} & 94.70  & 90.46   & 26.28& 5.34\\
{w/o-Pseudo} & 88.75 & 90.50   & 26.37&  5.60 \\  
w/o-Weight& 94.89& 89.30 & 25.65 &5.23\\
w/o-GRU& 93.76& 90.02& 25.44& 5.31 \\ \hline
{w-Fixed}  & 94.07  & 89.30 & 25.28 &  {5.24}  \\ 
{w-Learnable} & 96.05 & 90.94  & 25.94& 5.46  \\ \hline
{StyleD2T}  & \textbf{97.49}* & \textbf{93.44}* & \textbf{26.45}* & \textbf{5.77}* \\ \hline
\end{tabular}
}
\vspace{-0.5em}
\end{table}

\begin{figure}[!t]
  \centering
  \includegraphics[scale=0.53]{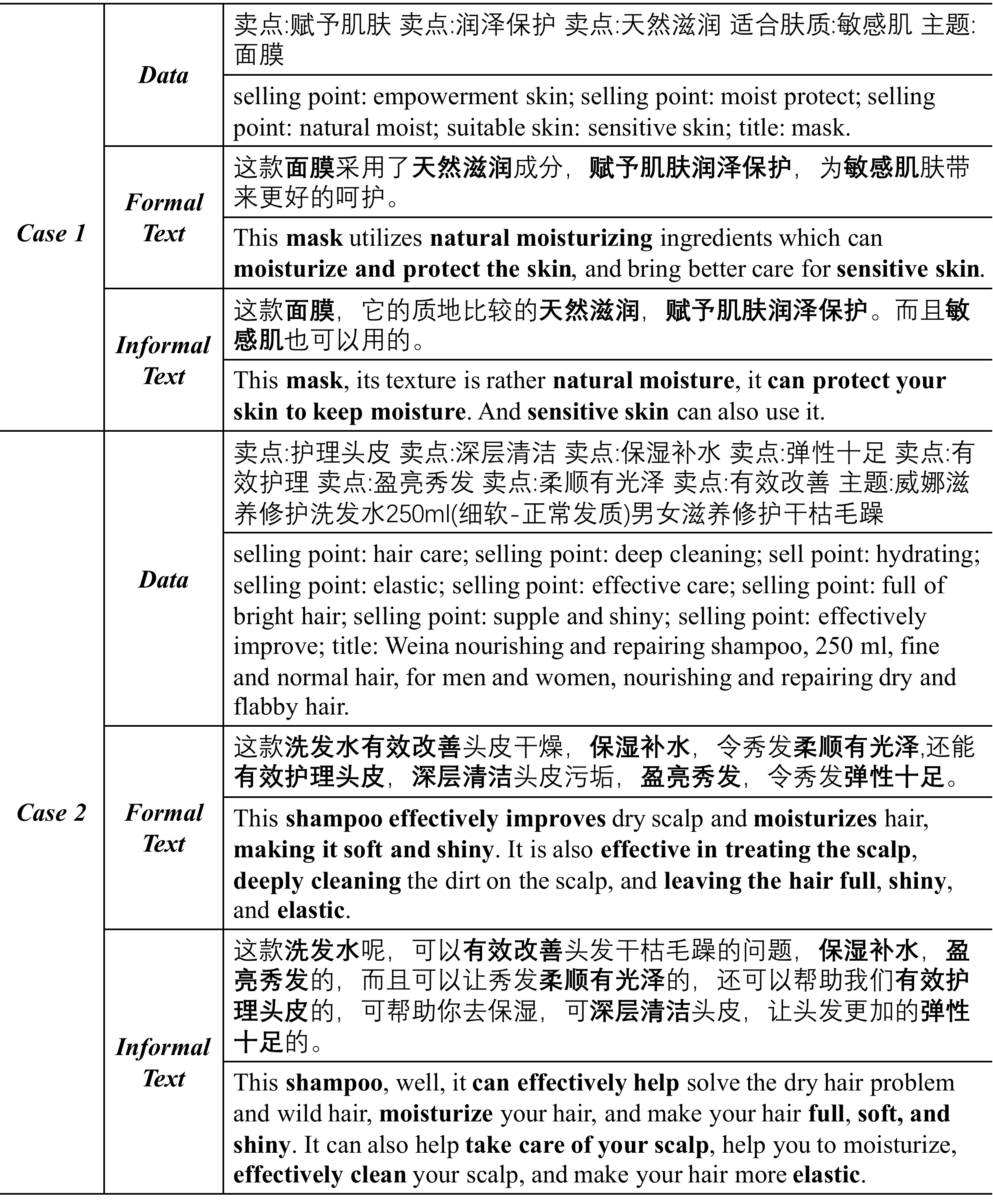}
  \caption{Generated cases of StyleD2T on the TaoStyle dataset. The English texts are translated from the Chinese texts, where some colloquial terms are difficult to translate.}
  \label{cases}
  \vspace{-1.0 em}
\end{figure}

\subsection{On Ablation Study (RQ2)}
To justify each module of our model and the superiority of stylized text input over using fixed or learnable vectors, we devise the following variants of StyleD2T.
\begin{itemize}
\item w/o-Style. In this variant, we discarded the mask-based style embedding by setting $\textbf{s}=AVG(\textbf{H}_X)$ in Eqn. ($\ref{eq11}$);

\item  w/o-StyleCon. In this variant, we removed the two constraints for style embedding by setting $\beta=0$ and $\gamma=0$;

\item w/o-Planner. In this variant, we disabled the logic planner by changing Eqn. ($\ref{planning}$) to $\mathbf{H}_o =Transformer_o([\textbf{E}^{1}, \cdots, \textbf{E}^{K}])$;

\item w/o-Graph. In this variant, we directly fed the original attribute-value pair embeddings of the data to the GRU rather than the enhanced ones;

\item  w/o-Pseudo. In this variant, we disabled the pseudo triplet augmentation by setting $\delta=0$.

\item w/o-Weight. \textcolor{black}{To explore the effect of weight in the logic graph, we set all the weights to $1$, \ie $e_{i,j}=1$ in Eqn. ($\ref{graph_cons}$).}

\item w/o-GRU. \textcolor{black}{To show the benefit of the GRU-based planning, we removed it by setting $\mathbf{H}_o=Transformer_o([\textbf{E}^{m};\cdots;\textbf{E}^{K}])$ in Eqn. ($\ref{planning}$)}

\item  w-Fixed. In this variant, we replaced the style embedding $\textbf{s}$ in our model with a fixed vector.

\item w-Learnable. Similar to w-Fixed, in this variant, we replaced the style embedding $\textbf{s}$ with a learnable vector.
\end{itemize}

\textcolor{black}{Table~\ref{RQ4} shows the ablation study results, from which we have made the following observations. 1) Removing any module of our model decreases its performance, which demonstrates the necessity of each module. 2) Removing the whole mask-based style embedding module (\ie w/o-Style) or only the two constraints for style embedding (\ie w/o-StyleCon) significantly decreases the style accuracy, implying the effectiveness of the feature-level mask design and the two constraints in style information extraction from unstructured style reference text. 3) Disabling the logic planner (\ie w/o-Planner) would largely decrease coverage, which suggests that a good planner can facilitate the subsequent target text generation. 4) Our model exceeds its variants w/o-Graph, w/o-Weight, and w/o-GRU, especially in the coverage metric, which suggests that logic graph, GRU-based planning, and the weight modeling of the logic graph are vital for the planner.  5) Our model substantially improves the style accuracy compared with w/o-Pseudo, which verifies its benefit in the unbiased stylized generation.  6) Our model outperforms w-Fixed and w-Learnable, demonstrating the advantage of using flexible style reference text to deliver the style information.}

\subsection{On Case Study (RQ3)}
We show two cases generated by our StyleD2T model in Figure~\ref{cases}. Based on the generated texts, our model can maintain the semantic consistency well for the given data. For example, in Case 1, the attribute values of the given data are all covered by the formal and informal texts that are generated. Meanwhile, we noticed that the texts generated by our StyleD2T model in these two cases meet the requirement of style consistency. For example, in Case 2, the formal text contains certain formal expressions, such as ``It is also effective in treating the scalp, deeply cleaning the dirt on the scalp, and leaving the hair full, shiny, and elastic'', while the informal text has some colloquial expressions, such as ``This shampoo, well, it can effectively help solve the dry hair problem and wild hair, moisturize your hair, and make your hair soft and shiny''. Overall, our StyleD2T model can generate coherent text for the given attribute-value pairs according to different styles.
   
\section{Conclusion and Future Work} \label{sec:conclusion}
In this paper, we define a new stylized data-to-text generation task and conduct a case study in the e-commerce domain. In particular, we present a novel framework, StyleD2T, which consists of three key components: logic planning-enhanced data embedding, mask-based style embedding, and unbiased stylized text generation. Meanwhile, we collect a real-world dataset named TaoStyle, which consists of 31,728 triplets, to evaluate the stylized data-to-text generation for formal and informal styles. Extensive experiments on the dataset demonstrate the superiority of our model over existing state-of-the-art methods and verify the effectiveness of each key module of our model. In particular, we find that the graph-guided logic planner helps to improve the coverage of the generated text; the mask-based style embedding indeed boosts the style information extraction, while the pseudo triplet augmentation benefits the unbiased text generation. 

In the future, we plan to explore more diverse input data forms, such as tables and English text. As a pioneer study, we also expect that more studies can be conducted in this research direction.


\begin{acks}
This work is supported by the National Natural Science Foundation of China, No.: U1936203; the Major Basic Research Project of Natural Science Foundation of Shandong Province, No.: ZR2021ZD15; the Shandong Provincial Natural Science Foundation, No.:ZR2022YQ59, and Alibaba Group through Alibaba Innovative Research Program.
\end{acks}

\bibliographystyle{ACM-Reference-Format}
\bibliography{reference}

\appendix

\end{document}